\documentclass[10pt,journal,compsoc]{IEEEtran}
\ifCLASSOPTIONcompsoc
  \usepackage[nocompress]{cite}
\else
  \usepackage{cite}
\fi
\ifCLASSINFOpdf
\else
\fi
\usepackage{float}
\usepackage{times}
\usepackage{epsfig}
\usepackage{graphicx}
\usepackage{amsmath}
\usepackage{amssymb}
\usepackage{subcaption}
\usepackage{rotating}
\usepackage{comment}
\usepackage{graphicx}
\usepackage{algorithm}
\usepackage{algorithmic}
\usepackage{hyperref}
\usepackage{url}
\usepackage{amssymb}
\usepackage{amsmath}
\usepackage{booktabs} 
\usepackage{algorithm}
\usepackage{algorithmic}
\usepackage{setspace}
\usepackage{comment}
\usepackage{hyperref}
\usepackage{graphicx}
\usepackage{algorithm}
\usepackage{algorithmic}
\usepackage{setspace}
\usepackage{comment}
\usepackage{microtype}
\usepackage{graphicx}
\usepackage{epsfig}
\usepackage{bbm}
\usepackage{epstopdf}
\usepackage{nicefrac}
\usepackage{amsmath,amsfonts,amsthm,bm}
\usepackage{booktabs, multicol, multirow}
\begin{document}

%
\title{Tasks Structure Regularization in Multi-Task Learning for Improving Facial Attribute Prediction}

%
%
%
%

\author{Fariborz Taherkhani, Ali Dabouei, Sobhan Soleymani, Jeremy Dawson, and Nasser M. Nasrabadi\\
Lane Department of Computer Science and Electrical Engineering \\ West Virginia University}

%
%

\markboth{Journal of \LaTeX\ Class Files,~Vol.~14, No.~8, August~2015}%
{Shell \MakeLowercase{\textit{et al.}}: Bare Demo of IEEEtran.cls for Biometrics Council Journals}
%



\IEEEtitleabstractindextext{%



\begin{abstract}
The great success of Convolutional Neural Networks (CNN) for facial attribute prediction relies on a large amount of labeled images. Facial image datasets are usually annotated by some commonly used attributes (e.g., gender), while labels for the other attributes (e.g., big nose) are limited which causes their prediction challenging. To address this problem, we use a new Multi-Task Learning (MTL) paradigm in which a facial attribute predictor uses the knowledge of other related attributes to obtain a better generalization performance. Here, we leverage MLT paradigm in two problem settings. First, it is assumed that the  structure of the tasks (e.g., grouping pattern of facial attributes) is known as a prior knowledge, and parameters of the tasks (i.e., predictors) within the same group are represented by a linear combination of a limited number of underlying basis tasks. Here, a sparsity constraint on the coefficients of this linear combination is also considered such that each task is represented in a more structured and simpler manner. Second, it is assumed that the structure of the tasks is unknown, and then structure and parameters of the tasks are learned jointly by using a Laplacian regularization framework. Our MTL methods are  compared with competing methods for facial attribute prediction to show its effectiveness.
\end{abstract}

}

\maketitle

\IEEEdisplaynontitleabstractindextext

%
\IEEEpeerreviewmaketitle

\IEEEraisesectionheading{\section{Introduction}\label{sec:introduction}}
\IEEEPARstart{F}{acial} attributes such as "blond hair" and "narrow eyes" are  soft biometrics which can be predicted from  face images. These biometrics are robust visual  features that can be used in a variety of applications such as face image retrieval \cite{taherkhani2018deep,talreja2018using}, face recognition \cite{taherkhani2018facial,talreja2019attribute} and face search engines \cite{hu2017attribute, kumar2008facetracer}. For example, Kumar et al. \cite{kumar2008facetracer} created a face search engine that users can simply query on statements such as "old man wearing glasses" to retrieve related images from a database. In this search engine, faces are ran through several attribute predictors which assign labels to each image in the database. Then, the related images from the database are retrieved based on the predicted attributes.

Recent developments in CNNs have provided promising results for many tasks in  biometrics such as facial attribute prediction \cite{zhong2016face, kalayeh2017improving,zhuang2018multi,sharma2020slim,mao2020deep}. However, the success of CNN models requires  a vast  amount  of well-annotated training images, which is not always feasible to perform manually \cite{NIPS2012_4824}. Moreover, most datasets annotated by facial attributes are not large enough, or images in the datasets are only  annotated by some commonly used facial attributes (e.g., gender) while label of the other facial attributes (e.g., big nose) are not available over the course of training. 

There are two different solutions that are usually used to improve the performance of a CNN model in such a case: 1) Semi-Supervised Learning (SSL), and 2) Transfer Learning (TL). In SSL methods \cite{chapelle2009semi, oliver2018realistic}, there is  a vast amount of unlabeled images but only a small amount of labeled images available over the course of training. As such, SSL methods tend to learn discriminative models that can make use of the information from an input distribution that is provided by a large amount of unlabeled images.  In TL methods \cite{pan2010survey, tan2018survey}, however, the learning of a new task is improved by transferring  knowledge from a related task which has already been learned, or the new task  is learned simultaneously with multiple prediction tasks that are related to it, the latter is also known as  Multi-Task Learning (MTL) paradigm. The main goal in MTL is essentially to share common information relevant to prediction among the tasks and learn them jointly to achieve a better generalization performance than learning each of them alone and independently. The common aspect in all MTL methods is to introduce an inductive bias in the joint hypothesis space of the related tasks which considers our prior beliefs about the structure or relatedness of the tasks. For example, assumptions such as  parameters of the related tasks share a common prior \cite{rai2010infinite},   parameters  of the related tasks lie near to each other on a manifold \cite{agarwal2010learning}, and  parameters of the related tasks lie in a low dimensional subspace \cite{kang2011learning} are examples of inductive biases which are used  for obtaining  a better generalization performance in MTL.

\begin{figure}[t] 
\centering
\includegraphics[width=8.8cm]{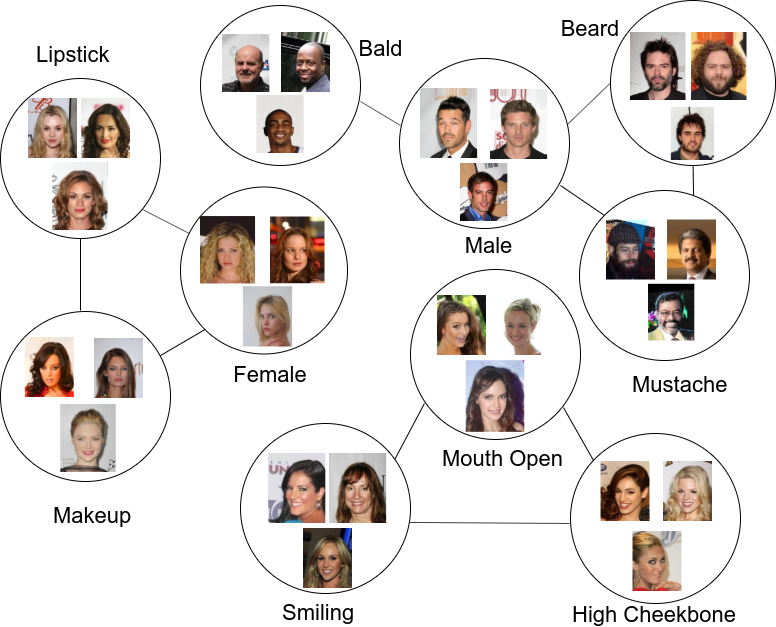}
\caption{Cluster structure of the facial attributes.}\label{fig:cmh_1}
\end{figure}

 It has been shown that facial attributes prediction task is not completely a de-coupled learning problem \cite{liu2015deep,ranjan2017all, zhuang2018multi,fan2019multi,hand2017attributes, mao2020deep}.  The rationale is  that the presence or absence of some facial attributes may suggest the presence or absence of some other facial attributes due to the strong correlation between facial attributes  that may appear simultaneously in the facial images. For example, facial attributes such as "mouth slightly open" and "smiling" quite possibly appear together in a face image. These related attributes can be predicted jointly via MTL methods \cite{ranjan2017all, han2017heterogeneous} by sharing relevant information to prediction among their predictors to improve the overall prediction accuracy.

 Although some facial attributes are related to each other, we would  expect that these facial attributes are related to each other to varying degrees which we usually do not know much about the strength of the relationships. Previously,  Liu et al. \cite{liu2015deep} indicated that the facial attributes have implicit grouping patterns which can be clustered into a few groups, with weak relationships among the tasks across the clusters, but strong relationships among the tasks within each cluster.  As an example shown in Fig. \ref{fig:cmh_1}, the facial attributes have been grouped into several clusters, where the attributes within a cluster have a strong relationship and those between the clusters have a weak relationship. In cases where structure of the tasks (i.e., level of relatedness) is provided as a prior knowledge during the training, this knowledge can then be encoded in a graph of tasks using a matrix which can be used as a relevant information to prediction in the MTL optimization problem \cite{evgeniou2005learning,argyriou2013learning}. However, structure of the tasks such as the one defined  in \cite{liu2015deep} may not be ideal, or  in the general case, structure of the tasks for the facial attributes may not be available as a prior knowledge during the training. In such a case, there is little option other than learning the structure and parameters of the tasks jointly to predict facial attributes in a MTL paradigm.

 In this work, we consider two scenarios for MT facial attributes prediction.  In the first scenario, the structure of the tasks is given as a prior knowledge  during the training. Here, we  use the term task for the facial attribute predictor. In this scenario,  we follow the model introduced in \cite{liu2015deep} which clusters facial attributes into a few groups. Although the level of dependency among facial attributes has not been specified in this model, the clustering pattern in this model can still provide useful knowledge for  optimizing  the parameters of the tasks in a MTL paradigm. For example, this clustering pattern can be used as an inductive bias in which the parameters of the tasks within the same cluster lie in a low dimensional subspace \cite{kang2011learning,daume2013learning,argyriou2008convex}. This is because, in this inductive bias, the level of tasks relatedness is not required. Following  previous works, we  assume that the  parameters of the tasks within the same cluster lie in a low dimensional subspace and also the parameters of each task is a linear combination of a limited number of basis tasks which construct the joint-hypothesis space of the related tasks. In this model, the coefficients of the linear combination are sparse in nature and the overlap in the sparsity patterns of two tasks indicates the amount of sharing knowledge across the related tasks. The sparsity constraint in our model is because whenever possible, a task is represented in a more structured and simpler manner.
 
In the second scenario, the structure of the tasks is unknown and we aim to learn the  parameters and structure of the tasks jointly based on the kernel methods \cite{evgeniou2005learning}. In other words, we formulate facial attribute prediction as an optimization problem in a reproducing kernel Hilbert space, in which the kernel are also learned. Therefore, our optimization problem is similar to the learning problem in an infinite set of kernels. However, our optimization problem differs as the feasibility set in our optimization problem is not convex where we use an alternating minimization algorithm to estimate the structure and parameters of the tasks. In this learning problem, structure of the tasks is learned through a graph in which each node represents a task, and weight of the edge between two nodes denotes the level relatedness between the two tasks. This graph is used as a relevant information to prediction in our MTL optimization problem to encourage the parameters of the related tasks to be similar to each other depending on the weight between the tasks in the graph. In other words, the distance between the parameters of the related tasks is regularized during the training such that their parameters become similar to each other in the joint-hypothesis space of the tasks.

 \subsection{Paper Organization}
 
 The rest of the paper is structured as follows: recent
related work to facial attribute prediction and MTL are briefly reviewed
in the next section. We  explain our general MLT framework and define facial attribute prediction problem in a MTL paradigm in Section 3. In Section 4, we introduce our inductive biases for multi-task facial attribute prediction, and then explain our methods for learning the parameters and structure of the facial attribute predictors in a MTL paradigm.   We experimentally evaluate our proposed multi-task model for facial attribute prediction on publicly  available  datasets annotated by facial attributes in Section 5. Finally, we conclude our research in Section 6.

\section{Related Work}
In this section, we briefly explain related work for facial attribute prediction, and MTL models, and also the work which have used MTL paradigm in deep learning models.  
\subsection{Facial Attribute Prediction}

Facial attribute prediction methods are roughly divided  into  local and global approaches. Local methods typically consist of three steps; first they detect different parts
of the object and then they extract features from each part. A combination of  these features is then
used to train a classifier \cite{kumar2009attribute,bourdev2011describing,chung2012deep,berg2013poof,zhang2014panda, moeini2017regression}. For example, Moeini et al. \cite{moeini2017regression}  propose a local method that assigns  a confidence level to each facial attribute indicating what percentage of a particular facial attribute is present in the image. In this method, facial landmarks are detected  using a multi-level feature representation obtained by constrained local model \cite{saragih2011deformable},  and finally, facial attributes are predicted through simultaneous dictionary learning.  The local methods usually work poorly if object localization and alignment are not ideal \cite{liu2015deep}. 

In contrast to local methods, global approaches \cite{liu2015deep,rudd2016moon,ehrlich2016facial,zhang2014panda,zhong2016face} ignore object parts or landmarks within the image and therefore, extract features from the entire image.  For example, Zhang et al. \cite{zhang2014panda} extract poselets based on deformable part models. The model aligns training samples globally for the entire object to cancel differences in the pose and view angle. This representation enables pooling across different poses and viewpoints to facilitate facial attribute prediction. The main downside of the global methods is lack of robustness to  objects deformations \cite{razavian2014cnn}.

Recently, the performance of attribute predictions has been improved by leveraging CNN models \cite{kalayeh2017improving, liu2015deep,zhong2016face,rudd2016moon,BMVC2016_131,zheng2020blan,sethi2018residual,duan2020novel,duan2020novel,sharma2020slim}. For example, Kalayeh et al. \cite{kalayeh2017improving} propose a part-based attribute prediction
method which deploys a semantic segmentation method based on a CNN model  to
transfer localization information from the auxiliary task of
semantic face parsing to the facial attribute prediction task.
Liu et al. \cite{liu2015deep} use two cascaded CNNs, the first of which,
LNet, is used for face localization, while the second, ANet,
is used for attribute description. Zhong et al. \cite{zhong2016face} first localize face images and then use an off-the-shelf architecture
designed for face recognition to describe face attributes at
different levels of a CNN. 

MTL framework for facial attribute prediction has been investigated in recent work \cite{liu2015deep,rudd2016moon,BMVC2016_131, ranjan2017all, zhuang2018multi,fan2019multi,hand2017attributes, mao2020deep}. For example, Rudd et al. \cite{rudd2016moon} demonstrated that multi-task optimization is better for CNN-based models to predict facial attributes. They introduce a novel Mixed Objective Optimization Network (MOON) which uses a loss function mixing multiple task objectives with domain adaptive re-weighting of propagated loss.  He et al. \cite{BMVC2016_131} also propose a MTL framework for relative attribute prediction. The method
uses a CNN to learn local context and global style information from the intermediate convolutional and fully connected
layers, respectively. 
Moreover, Ehrlich et al. \cite{ehrlich2016facial} introduce a MTL model based on a restricted Boltzmann machine to create a shared feature representation for multiple attribute prediction. This model is performed on faces and facial landmark points directly to learn a shared feature representation over all existing facial attributes. In other case, Sethi et al. \cite{sethi2018residual} propose a model utilizing a cosine similarity-based loss function in an auto-encoder to represent input face images for facial attribute prediction task.

\subsection{Multi Task Learning}

MTL  have provided promising results for computer vision and biometrics tasks, especially for the tasks where only a few  samples are available over  the course of training \cite{pan2010survey, tan2018survey}. In MTL, the knowledge learned from one task is transferred to the other tasks for training. It has been shown that  knowledge sharing can increase performance of some, or occasionally  all tasks \cite{pan2010survey,weiss2016survey}.

Many approaches in the context of MTL have been proposed to share knowledge between tasks. These approaches, depending on the relationship between tasks or features, are roughly divided into three categories. In the first category,  the relationship between the tasks are incorporated in the training phase. For example, Hariharan et al. \cite{hariharan2010large} formulate the MTL problem as a multi-label classification problem with a prior knowledge about densely correlated labels. The model introduces a max-margin multi-label formulation and includes correlation-based interactions between labels in the prediction loss function. The methods in the second category usually discover a common feature structure between all the tasks or explore  common related features between them. For example, Rai et al. \cite{rai2010infinite} propose a non-parametric Bayesian model to learn a latent shared subspace for all the related tasks. The shared subspace embeds the relationships between all the tasks in a way that the parameters of each task (for example the weight vector in  linear SVM) are interpreted  as a linear combination of a set of basis tasks that form this latent shared subspace \cite{daume2013learning,kang2011learning}. The third category of approaches, which are the recent models, such as max-margin method \cite{zhang2010learning} and Bayesian method \cite{yang2013multi}, consider both the tasks and feature correlations. Even though max-margin method is stronger in discrimination; Bayesian method is more flexible on incorporating rich prior inference \cite{li2014bayesian}.

\subsection{Multi Task Learning in Deep Models}

MTL in deep models is usually performed  by either hard or soft parameter sharing  \cite{long2017learning,lu2016fully,misra2016cross,yang2016deep}. Hard parameter sharing is performed by sharing the convolutional layers between all tasks, and then branching out to several fully connected layers in which each branch is dedicated to a particular task. This sharing strategy considerably diminishes the risk of overfitting. In models based on hard parameters sharing, it  has been shown that if the number of  tasks which are learned simultaneously is increased, the model must find a representation that captures all the tasks \cite{long2017learning, lu2016fully}. This results in reducing the chance of overfitting on the original task. In soft parameter sharing, however, every task has a separate model and parameters. In models based on soft parameters sharing, the distance between the parameters of the related tasks is regularized during the training step such that their parameters are encouraged to be similar to each other. These models usually use a regularization framework by introducing an inductive bias to avoid overfitting.

Some of recent efforts based on hard and soft parameters sharing strategies are \cite{long2017learning,lu2016fully,misra2016cross}. For example, Long et al. \cite{long2017learning} enhance MTL in deep models by proposing Deep Relationship Networks (DRN). The DRN  not only shares network structure and task-specific layers, it also puts a prior matrix,  similar to Bayesian models on the fully connected layers \cite{rai2010infinite, yang2013multi}, to learn relationships between the tasks. This network still requires a pre-defined structure to share the parameters, while being sufficient for simple tasks, is not optimal for difficult tasks. Lu et al. \cite{lu2016fully} propose a bottom-up approach with a fully-adaptive feature sharing strategy. The model  starts with a slim network, and is dynamically widened in a greedy fashion so that similar tasks are grouped together during the training. The widening procedure creates branches dynamically. The downside of this approach is that the greedy method sometimes can not construct a model which is globally optimal as in this method, each branch is assigned  to only one task and the model does not have the chance to learn more complex interactions between the tasks. As an example of soft parameter sharing models, Misra et al. \cite{misra2016cross}  propose a model which starts with two separate networks, and then place units called cross-stitch units only after pooling and fully-connected layers to enable the task-specific networks to use the knowledge of other related tasks. This process is performed by learning a linear combination of the previous layers output. 

\section{Tasks Structure Regularization in MTL for Facial Attribute Prediction}

\begin{figure}[t]
\centering
\includegraphics[scale=0.385]{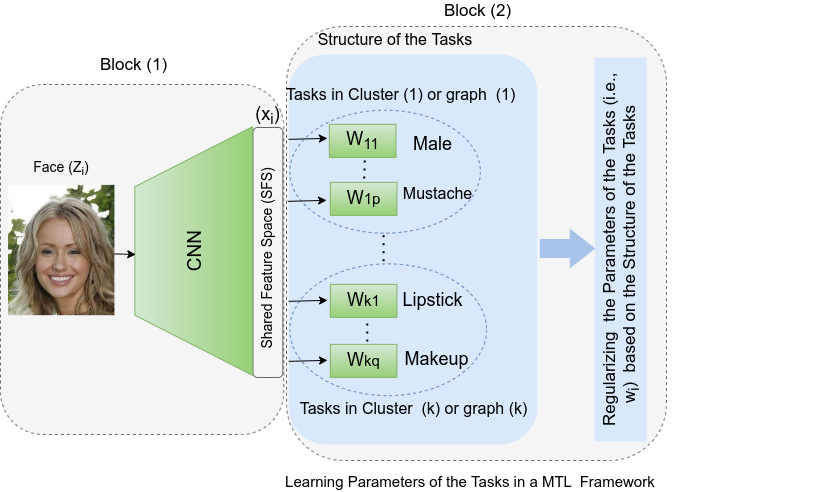}
\caption{Multi-Task-Facial Attribute Prediction (MTFAP) : General framework for regularizing the parameters of the facial attribute predictors in a MTL paradigm.}
\label{arc1}
\end{figure}

Before formulating the tasks structure regularization in MTL for facial attribute prediction, we first explain our overall MTL schema which is shown in Fig. \ref{arc1}  for joint-facial attribute prediction. We call this framework Multi-Task Facial Attribute Predictor (MTFAP).  MTFAP contains two blocks as is shown in Fig . \ref{arc1}. In block (1), we  simply use  a CNN model.  The CNN in our MTFAP shares the convolutional layers between all the tasks (i.e., facial attribute predictors), and then fan out to several fully connected layers in block (2), each of which is dedicated to predict a particular facial attribute. Likewise most of the deep MTL models \cite{long2017learning,lu2016fully,misra2016cross,yang2016deep}, MTFAP also shares the general visual patterns in a common Latent Feature Space (LFS in Fig. \ref{arc1}) among all the tasks and then learns more localized features for each specific task when all the  tasks are trained jointly. In our MTFAP, parameters of the predictors are regularized based on the structure of the tasks which is represented by a graph or a group of clusters to predict facial attributes in a MTL fashion. In this work, we are naive in designing block (1), and we use a generic CNN model and we mostly focus on block (2) to design an effective regularization schema to improve the prediction accuracy.

 Here, we formulate tasks structure regularization in MTL for facial attribute prediction. Here, the input to our CNN, as is shown in Fig. \ref{arc1}, is an image and the output is a vector in which each of the element indicates the presence or absence of a  particular facial attribute. In our formulation, $\mathcal{D}=\{{z}_1,...,{z}_n\}$ denotes images in the training set, $\mathcal{X}=\{x_1,...,x_n\}$ denotes d-dimensional features extracted by the CNN from the training images, and $\mathcal{Y}_j=\{{y}_{j1},...,{y}_{jn}\}$ represents labels of the \textit{j-th}  attribute for the training images where  ${y}_{ji}$ is a binary value indicating whether the \textit{j-th} facial attribute is present in the image $z_i$ or not.

 
Assume that there are \textit{k} tasks (i.e., facial attribute predictors), and  let $\mathcal{L}(\mathcal{X},\mathcal{Y}_j, w_j)$ denote the loss function for learning the \textit{j-th} task on the training images $\mathcal{D}$, and $w_j \in \mathbb{R}^d$ represent parameters of the \textit{j-th} task, and $W \in \mathbb{R}^{d\times k}$ be a matrix constructed by stacking the parameters of $k$ tasks (i.e., $[w_1,...,w_k]$) column-wise. In the general and standard learning paradigm, tasks are learned independently, and no information are shared among the other tasks. In this learning paradigm, parameters of a given task are usually penalized by using an $\ell_2$ norm (i.e., $ \sum _{i=1}^k||{w}_i||_2$)  and  dependency or coupling between the task and other related tasks are ignored. In other words,  all the tasks are learned separately and there is no interaction among the related tasks in the training loss function to share the knowledge. Thus, the loss function for learning the parameters of $k$ tasks, $W$ is formulated as:

\begin{equation}
W=\operatorname*{argmin} \sum_{i=1}^k \mathcal{L}(\mathcal{X},\mathcal{Y}_{i}, {w}_i)+ \lambda ||w_i||_2, 
\end{equation}
where $\lambda$ is a hyperparameter that creates a balance between two terms in the loss function.  However, in MTL framework, the main goal is to consider dependencies  between the related tasks  and then regularize the parameters of the tasks during the training. To regularize the parameters of the tasks, the most important step is to introduce  an inductive bias in the joint-hypothesis space of the all tasks which considers our prior beliefs about the structure of the tasks. For example, as an inductive bias, we can force parameters of the related tasks to be similar to each other. This inductive bias is used in MTL based on the soft parameter sharing methods which aim to regularize the distances between the parameters of the related tasks during the training such that their parameters become similar  to  each  other in the joint-hypothesis space of the all tasks \cite{long2017learning,lu2016fully,misra2016cross,yang2016deep}. Generally, the inductive bias is applied by using a function operating on the matrix of tasks parameters (e.g., $\Omega (W)$) which is integrated in the total training loss function. Thus, the loss function for learning \textit{k} related tasks in a  MTL paradigm is formulated as:

\begin{equation}
W=\operatorname*{argmin} \sum_{i=1}^k \mathcal{L}(\mathcal{X},\mathcal{Y}_{i}, {w}_i)+ \lambda \Omega(W).
\end{equation}

In this work, we are naive  on $\mathcal{L(.)}$ loss term, and we use the generic cross-entropy loss function, but we  mostly focus on the inductive bias which is used in the regularization term (i.e., $\Omega (W)$) in our total loss function to learn the parameters of $k$ tasks in a MTL paradigm. In our MTL framework, after performing many none-linear operations on the input images using the baseline CNN,  each facial attribute predictor (e.g., $w_i$) follows a linear model (i.e., $f_i(x_j)={w}_i^\top x_j$) to predict the presence of the \textit{i-th} attribute in the image ${z}_j$, where ${x}_j$ is the output of the SFS layer for the image ${z}_j$.

\section{Our Inductive Biases for Regularizing Structure of the  Tasks in MTFAP}
As discussed in the introduction and section 3, the
critical aspect in MTL approaches is
to introduce a proper inductive bias in the joint-hypothesis space of the tasks to consider our prior beliefs about  their structure. This inductive bias generally is reflected with a regularization term (e.g., $\Omega(W)$) in the total training loss function as is formulated in Eq. (2).

 In this work, we consider two scenarios for  multi-task facial attribute prediction.  In the first scenario, we assume that structure of  the tasks is given as a prior knowledge during the training. In this scenario, our prior knowledge is that facial attribute predictors  are grouped into several clusters,  where  the  predictors  within  the same  cluster  have  strong relationships  and  those  across  the clusters  have  weak relationships, or in some cases there is no relationship between them at all. In this scenario, following the other MTL methods \cite{agarwal2010learning,kang2011learning,srebro2005maximum,kumar2012learning, ando2005framework,argyriou2008convex}, we use an inductive bias where  parameters of the tasks (i.e., predictors)  within  the same  cluster  lie  in  a  low dimensional  subspace and then parameters of each task  is represented by a  linear  combination  of  a  finite number of  basis  tasks which  construct  the  joint-hypothesis space of these tasks.  Moreover, we also assume that the coefficients of this linear combination are sparse in nature and the overlap in the sparsity patterns of two tasks indicates the amount of sharing knowledge among them. The sparsity constraint in this setting is because whenever possible, a task is represented in a more structured and simpler manner \cite{srebro2005maximum,kumar2012learning}.

 In contrast to the first scenario, in the second scenario,  structure of the tasks is unknown and we aim to learn the parameters and  structure of the tasks jointly from the training data. In this scenario, we assume that attribute predictors are related with each other to varying degrees, some of them are strongly related while others are weakly related. This assumption motivates us to encode the structure of the tasks in a graph in which, each of the node in the graph represents an attribute predictor and the weight of the edge between two attribute predictors represents the strength of their relationship. Then, the goal is to incorporate this graph in our MTL framework such that strongly related attribute predictors share  more knowledge  from  the  SFS  layer (in Fig. \ref{arc1}) while allowing the unrelated attribute predictors are trained independently without  influencing  on  each  other,  and consequently  share  no knowledge from SFS layer for the prediction. In other words, we encourage the parameters of two attribute predictors to lie close to each other in some geometric sense, depending on the strength of the relationship between them in the graph.
 
 In the following subsections, we explain training of our multi-task facial attribute prediction using the inductive biases described earlier for these two scenarios. 
 \subsection{Multi-Task Facial Attribute Prediction with Tasks Structure as a prior knowledge}
 In this section, we explain our MTL framework for facial attribute prediction when structure of  the tasks is available as a prior knowledge during the training. In our work, we use structure of the facial attribute predictors introduced in  Liu et al. \cite{liu2015deep}. In this work, it is indicated that the facial attributes have implicit grouping patterns which can be clustered into a few groups, with weak relationships across the clusters, but strong relationships within each cluster.  As shown in Fig. \ref{threl_4}, the facial attributes have been grouped into several clusters, where the facial attributes in the same cluster have high level of relationships while facial attributes across the clusters have low level of relationships.

\begin{figure}[t]
\centering
\includegraphics[scale=0.3]{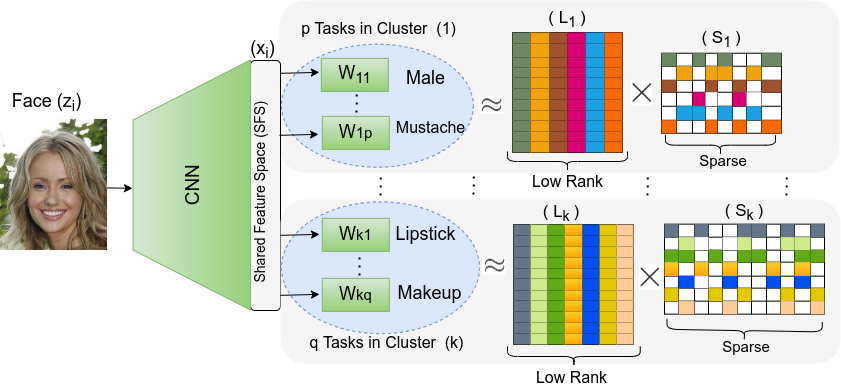}
\caption{Diagram of the first scenario: each task in the cluster is represented by the sparse combination of the basis tasks which construct the joint-hypothesis space.}
\label{arc2}
\end{figure}

Assume that there are \textit{m} groups of tasks, and in the \textit{g-th} group, there are $T_g$ tasks. Moreover, let $w_{gj} \in \mathbb{R}^{d}$ represent the parameters of the task in the \textit{g-th} group indexed by $j$, and also let ${W}_g=[w_{g1},w_{g2},...,w_{gT_g}] \in \mathbb{R}^{d \times T_g}$  be the matrix of the tasks parameters created by stacking the parameters of the each task within the \textit{g-th} group column-wise. Moreover, assume that for each group $g$,  there are $(K_g < T_g )$ latent basis tasks in the joint-hypothesis in which the parameters of the each task in the \textit{g-th} group can be represented by a linear combination of a subset of these basis tasks. This assumption stems from an inductive bias in the MTL where it is assumed that the tasks within  the same cluster  lie  in  a  low  dimensional  subspace \cite{kang2011learning,daume2013learning,argyriou2008convex}. Considering this,  we can then formulate the  matrix ${W}_g$ using ${W}_g = L_g \times S_g$, where $L_g \in \mathbb{R}^{d \times K_g}$ is a matrix in which each column represents a latent task, and $S_g \in \mathbb{R}^{K_g \times T_g}$ is a matrix which includes coefficients of the linear combination for each task in the \textit{g-th} group. Thus, $w_{gj}$ for the task $j$ in the \textit{g-th} group is obtained by
$L_g \times S_{gj}$, where $S_{gj}$ is the \textit{j-th} column of the matrix $S_g$. In our model, we put a sparsity constraint on the matrix $S_g$. This constraint forces each of the task in the \textit{g-th} group to be obtained by only a few number of the latent tasks which are indexed by non-zero elements of the corresponding column of the matrix $S_g$. This constraint is mainly because whenever possible, a task is represented in a more structured and simpler manner. Thus, considering $S_{gj}$ representing  sparse coefficients for the task $j$ in the \textit{g-th} group, and the tasks within  the same  cluster  lie  in  a  low  dimensional subspace, our total MTL loss function is then formulated as follows:

\begin{equation}
\begin{gathered}
\mathcal{L}(\mathcal{X},\mathcal{Y},L,S)=\\
 \sum_{g=1}^m \left(\sum_{j=1}^{T_g}\mathcal{L}_c(\mathcal{X},\mathcal{Y}_{gi}, L_g \times S_{gj})+ \lambda ||S_g||_1+\gamma ||L_g||_*\right), 
 \end{gathered}
\end{equation}

where $\mathcal{L}_c(.)$ is the cross-entropy loss function used for the prediction, $\mathcal{Y}_{gj}$ denotes the set of ground truth labels corresponding to the task $j$ in the $g$-th group, and $w_{gj}=L_g  \times S_{gj}$ is the weight vector or parameters of the task. In this model, the predicted label for the input $x_i$ is obtained by $(L_g \times S_{gj})^\top x_i$. Moreover, $||.||_*$ is the nuclear norm, $||.||_1$
is entry-wise $\ell_1$ norm,  and $\gamma$ and $\lambda$ are the hyperparameters that create a balance between terms used in our total loss function. The role of nuclear norm is to force a low rank penalty on the matrix $L_g$, which causes the parameters of the tasks within the same group lie in a low dimensional subspace. Originally, matrix $L_g$ should be penalized by the rank of the matrix. However, rank minimization is doubly exponential. Then, we relax it with the nuclear norm which is the tightest convex relaxation to the rank. Similarly, for sparsity constraint, we penalize the matrix $S_g$ with the $\ell_1$ norm instead of using the $\ell_0$ norm as the $\ell_0$ is a combinatorial norm, and we relax it with an $\ell_1$ norm \cite{candes2009exact}. Fig. \ref{arc2} graphically indicates the representation of  the tasks parameters via sparse coefficients of the basics tasks. Here, each color in the low rank matrix shows the parameters of a basic task, and  corresponding color in the sparse matrix denotes the contribution of the basic task in reconstruction of a  task parameters in the cluster. 

\subsubsection{Optimization Framework}
To obtain the parameter of the tasks in Eq. (3), $L=[L_1,L_2,...,L_m]$ and $S=[S_1,S_2,...,S_m]$ should be optimized during the training. By considering that the training images are projected into the shared latent feature space (i.e., SFS in Fig. \ref{arc2}), if the loss function defined in Eq. (3) is separately applied  on the extracted features, the optimization problem defined in Eq. (3) for a convex empirical loss function $\mathcal{L}_c (.)$, such as the cross-entropy is convex when one of  $L$ or $S$ is fixed. Thus, we  can then adopt an alternative minimization strategy to minimize the loss function.

\textbf{Fixing $S$ and optimizing $L$:} For the case where we fix $S$, we need to optimize Eq. (3) with respect to $L$. For this case, since the optimization problem contains a  nuclear norm and this norm is non-differentiable, there are basically better solutions based on the semi-definite programming or proximal gradients than the gradient-decent approach. However, the CNN is  usually trained by stochastic gradient descent. Therefore, we need to follow the standard training procedure and we then choose the sub-gradient schema for nuclear norm optimization \cite{yang2016trace}. The sub-gradient for the nuclear norm is $\frac{\partial||L||_*}{\partial||L||}=L(L^\top L)^{-\frac{1}{2}}$. Note that in this equation, in contrast to directly computing the inverse matrix square root, a more numerical stable solution 
is  $L(L^\top L)^{-\frac{1}{2}}=UV^\top$ in which $U$ and $V$ are computed by SVD as $L=U \Sigma V^\top$ \cite{watson1992characterization}. Thus, sub-gradient of a loss function containing a nuclear norm (e.g., $F(L)=f(L)+ \alpha ||L||_*$)  can be derived as follows: $\mathcal{G}(F(L)) \stackrel{\text{def}}{=} \frac{\partial f(L)}{\partial L}+ \alpha U V^\top$, where $f(L)$ can be any convex function but not necessarily differentiable \cite{avron2012efficient}.

 Algorithm 1 is a fast  Stochastic Sub-Gradient Descent (SSGD) algorithm based on QR factorization of SVD for updating the parameter $L$ \cite{avron2012efficient}.  In this Algorithm (i.e., line 2),  $Y$ is a probing matrix. A random $n \times k$ matrix $Y$ is considered as a probing matrix if $\mathbb{E}[Y Y ^{\top}] = I_{n \times n}$ where $I_{n \times n}$ is the $n \times n$ identity matrix and the expectation in this equation is over the choice of $Y$ \cite{avron2012efficient}. There are three different ways in \cite{avron2012efficient} to generate a probing matrix. However, we use the one in which $Y$ is a scaled identity matrix, because computation of $\mathcal{G}(F(X))Y$ is more efficient in such a case. Let $Y=Z/k$ where $Z$ is a random matrix created such that each column of $Z$ is drawn uniformly at random and independent of each other from a set of scaled identity vectors $\{\sqrt {n} e_1,...,\sqrt {n} e_n\}$. The variable $\eta^{(t)}$ as an input to the algorithm is the step size which the experimental results in \cite{avron2012efficient} has shown that using a fixed step size $\eta$ provides  better results. Thus, following \cite{avron2012efficient},  to make the step size scale free,  we set $\eta=\nu||Z||_F^2$ where $\nu$ is a parameter which the best value for it is around $0.009$.

\begin{algorithm}[t]
\caption{Fast SSGD-Update}
\begin{algorithmic}[1]
\label{algo1}
\STATE {\textbf{input}:  $U \in \mathbb{R}^{m \times r^{(t)}}$, $\Sigma \in \mathbb{R}^{r^{(t)}\times r^{(t)}}$, $V \in \mathbb{R}^{n \times r^{(t)}}$, $\eta^{(t)}$ }
\STATE $S^{(t)} \leftarrow{\mathcal{G}(\mathcal{L}(\mathcal{X}, \mathcal{Y},L,S))Y},$
\STATE $\hat{U}^{(t+1)} \leftarrow{[U^{(t)}\Sigma^{(t)}S^{(t)}]},$
\STATE $\hat{V}^{(t+1)}\leftarrow{[V^{(t)}-\eta^{(t)}Y]},$
\STATE Factorize: $\hat{U}^{(t+1)}=Q_U R_U,$
\STATE Factorize: $\hat{V}^{(t+1)}=Q_V R_V,$
\STATE $T\leftarrow{R_U R_V^{\top}},$
\STATE  Computing SVD: $T=M \bar{\Sigma}^{t+1}N^\top,$
\STATE $\bar{U}^{(t+1)}\leftarrow{Q_UM},$
\STATE $\bar{V}^{(t+1)}\leftarrow{Q_VN},$
\STATE $U^{(t+1)}\leftarrow{\bar{U}^{(t+1)}}$, $V^{(t+1)}\leftarrow{\bar{V}^{(t+1)}},$
\RETURN $\bar{U}^{(t+1)}$, $\bar{\Sigma}^{t+1}$, $\bar{V}^{(t+1)}$
\end{algorithmic}
\end{algorithm}

\textbf{Fixing $L$ and optimizing $S$:} For the case where $L$ is fixed, we need to optimize Eq. (3) with respect to $S$. Similar to the previous case, we update  $S$ in an online manner. Here someone may think of using the stochastic gradient descent method. However, unfortunately, the stochastic gradient descent approach fails to provide a sparse solutions, which also makes the $\ell_1$ minimization both slower and less practical as sparsity is one of the main reasons to use $\ell_1$ regularization \cite{shalev2011stochastic}. To solve this problem, several methods have been proposed \cite{shalev2011stochastic, duchi2008efficient,langford2009sparse}.  In this work, due to speed and facility for the implementation,  we use the method in  \cite{shalev2011stochastic} which is called Stochastic Mirror Descent Algorithm mAde Sparse (SMIDAS) to provide a proper sparse solutions for the $\ell_1$ regularized loss minimization. This optimization approach maintains two weight vectors: primal $s$ and dual $\theta$. Here, we denote a given column of matrix $S$ with $s$. The connection between the two vectors is made by a link function as follows: $\theta=f(s)$, where $f:\mathrm{R}^d\rightarrow\mathrm{R}^d$. Since the link function is  essentially considered as the gradient map of some strictly convex function $f$,  it is  invertible and  can  be written as follow: $s=f^{-1}(\theta)$. Following \cite{shalev2011stochastic}, we  also use the $p$-norm link function in which the \textit{j-th} element of $f$ is as obtained follows:

\begin{equation}
f_j(s)=\frac{sign(s_j)|s_j|^{q-1}}{||s||_q^{q-2}},
\end{equation}
where $||s||_q=(\sum_j|s_j|^q)^{\frac{1}{q}}$. We note that the function $f$ simply is  the gradient of  $\frac{1}{2}||s||_q^2$. Therefore, the inverse function for $f$ is obtained as follows \cite{gentile2003robustness, shalev2011stochastic}:

\begin{equation}
f^{-1}_j(\theta)=\frac{sign(\theta_j)|\theta_j|^{p-1}}{||\theta||_p^{p-2}},
\end{equation}
where, $p=q/(q-1)$. Algorithm 2 indicates one iteration for updating each of the column $s$ in the matrix $S$ using SMIDAS. Therefore, we need to repeat this procedure for all the columns  to obtain a sparse solution for $S$. In this Algorithm, $d$ and $r$ are the total number of the features, and step size, respectively. The parameter $r$ is usually a small number which we set it to $10^{-3}$.

\begin{algorithm}[t]
\caption{SMIDAS-Update}
\begin{algorithmic}[1]
\label{algo1}
\STATE {\textbf{input}:  $r >0$, $\lambda$ in Eq (3), $p=2 \ln(d)$, $f^{-1}_j(\theta)=\frac{sign(\theta_j)|\theta_j|^{p-1}}{||\theta||_p^{p-2}}$}
\STATE $v=\frac{\partial \mathcal{L}(\mathcal{X},\mathcal{Y},L,S)}{\partial s}$,
\STATE $\hat{\theta} =\theta-r v$,
\STATE $\forall j, \theta_j = sign (\hat{\theta}_j) max \{0,|\hat{\theta}_j|- r \lambda \}$,
\RETURN $s=f^{-1}(\theta) \Rightarrow s_j= \frac{sign(\theta_j)|\theta_j|^{p-1}}{||\theta||_p^{p-2}}$,
\end{algorithmic}
\end{algorithm}

\subsection{Multi-Task Facial Attribute Prediction without prior Knowledge on Tasks Structure}
In this section, we explain our MTL framework for facial attribute prediction where task relatedness structure is unknown. In our MTL model for this scenario,  we presume that tasks (i.e., attribute  predictors)  are  related  with  each  other  to varying degrees, some of them are strongly related while the others are  weakly related or in some cases there is no relationship between the tasks.  Therefore, this criterion allows us to encode the tasks relatedness structure using a graph in which, each of the node in the graph represents a task and the weight of an edge between two tasks represents the strength of their relationship. Leveraging the graph of task relatedness as a relevant information in the MTL optimization problem has been previously studied in \cite{evgeniou2005learning, sheldon2008graphical,argyriou2013learning, liu2018multi}. In these works, the goal is to penalize the tasks via a Laplacian quadratic form in the MTL optimization objective. Specifically, given a graph as a matrix $A_{m \times m}$, the Laplacian $L$ of this graph is derived as $L=D-A$, where $D=diag (d)$ is the diagonal matrix created by the degrees of the nodes $d_i=\sum_{j=1}^m A_{ij}$. The Laplacian regularization schema involved in the  MTL optimization loss function is written as follows:
\begin{equation}
\mathcal{L}(W,\mathcal{X},\mathcal{Y})= min \sum_{i=1}^n \mathcal{L}_c(\mathcal{X},\mathcal{Y}_{i}, {w}_i)+ \lambda \sum_{j=1}^n \sum_{k=1}^n L_{jk} w_j^\top w_k, 
\end{equation}
where $\lambda$ is the balancing term, and the first term is the loss term used for training each task independently while the second term is the Laplacian regularization term used to encourage the parameters of two tasks (i.e., $w_j$, $w_k$) to lie close to each other in some geometric sense, depending on $L_{jk}$ which is the strength of the relationship among them. This is essentially because the Laplacian regularization term equals $\sum_{j=1}^n\sum_{k=1}^n A_{jk} ||w_j-w_k||^2$ and thus this term in the loss function encourages the pairs of tasks ($w_j$, $w_k$) with the larger weights $A_{jk}$ to become  more similar.
In the next section, we explain about the Laplacian regularization framework for joint-attribute prediction and also, we formulate the optimization problem to jointly learn both the parameters of the tasks and the graph of the tasks relatedness.
\begin{figure}[t]
\centering
\includegraphics[scale=0.17]{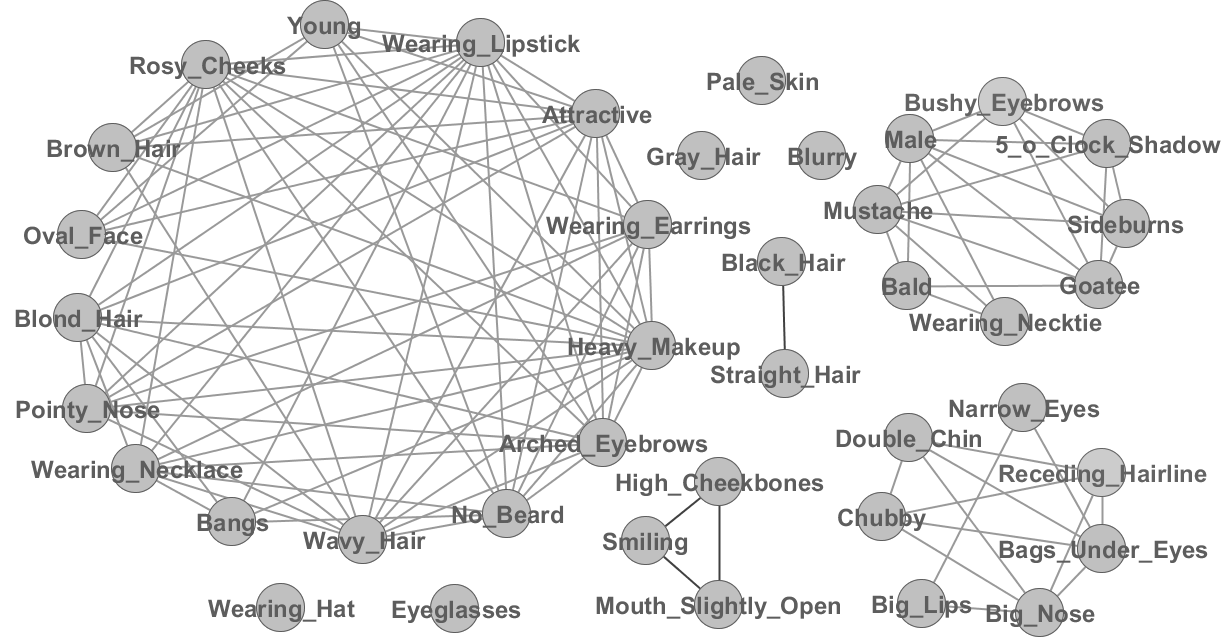}
\caption{Facial attributes relatedness encoded in the graph.}
\label{threl_44}
\end{figure}
\subsubsection{Joint-learning framework for the parameters and graph of the facial attribute predictors}

The goal in the MTL objective defined in (6) is  to consider coupling between tasks (i.e., facial attribute predictors). In other words, this objective function is a regularization approach which couples related tasks during the training process. If we just simply use an $\ell_2$ norm term, $ \sum _{i=1}^n||w_i||_2$, as a penalization term in the second term of (6), we will ignore the dependencies and coupling between related tasks. In such a case, we  learn all the tasks separately and there is no interaction between the related tasks in the penalization function to share the knowledge between the related tasks. In other words, in this MTL framework, we look to design a regularization function to satisfy two conditions. First, we should penalize each task separately; it means that each task should be simple enough.
Second, each task should be close to  another task in some geometric sense depending on a weight which indicates how much the two tasks are related to each other. For example, if  two given tasks are highly related to each other, these two tasks should be close to each other and consequently $ ||w_i-w_j||_2$ should be small in the regularization function. As  mentioned earlier, task relatedness among the tasks can be represented by a graph (see Fig. \ref{threl_44}).

By considering the two aforementioned conditions, we can formulate a regularization function to satisfy these two conditions as follows:
 Let $ M \in {\rm I\!R}^{n\times n}$ 
 be the matrix encoding the graph of task relations where $M_{ij} \geq 0$ indicating the level of relatedness between task \textit{i} and \textit{j}; then our regularization function can be written as follows:
 \begin{equation}
 \mathcal{L}_r (w_1, w_2,...w_n)=\sum_{i=1}^{n} \sum_{j=1}^{n} M_{ij}||w_i-w_j||^2 +\epsilon \sum_{q=1}^n ||w_q||^2.
 \end{equation}
 The first term in (7) satisfies the second condition of our  penalization function. It enforces related tasks to become closer to each other.  The coupling between the related tasks takes place by using this term,  and consequently, this term enforces the related tasks to share  more  visual  knowledge from the shared feature domain. The second term in (7)  satisfies the first condition for our penalization function, indicating that each task is penalized separately. The first term  can also be rewritten as follows:
 \begin{equation}
      \sum_{i=1}^{n} \sum_{j=1}^{n} M_{ij}||w_i-w_j||^2=\sum_{i=1}^{n} \sum_{j=1}^{n} L_{ij} w_i^\top w_j, 
 \end{equation}
 where $L$ is the graph Laplacian of task relatedness graph encoded by the matrix $M$:
\begin{equation}
    L=D-M,
\end{equation}
\begin{equation}
     D=diag(\sum_{j=1}^n M_{1j},...,\sum_{j=1}^n M_{nj}).
\end{equation}
Now we can rewrite, $\mathcal{L}_r (w_1, w_2,...w_n)$  as follows:
\begin{equation}
\mathcal{L}_r (w_1, w_2,...w_n)= \sum_{i=1}^{n} \sum_{j=1}^{n} A_{ij} w^\top_i w_j,
\end{equation}

where, \begin{equation}
  A=L+\epsilon I_{n \times n}. 
\end{equation}
It turns out that $\mathcal{L}_r (w_1, w_2,...w_n)=\mathrm{Tr}(W A W^\top)$ because we can expand it  as follows:
\begin{equation}
\begin{split}
   \mathrm{Tr}(W A W^\top)=&\sum_{i=1}^d W_i^\top A W_i= \sum_{i=1}^d\sum_{r,s=1}^n A_{rs} W_{ir} W_{is}
\\&=\sum_{r,s=1}^n A_{rs} \sum_{i=1}^d W_{is} W_{ir}=\sum_{r,s=1}^n A_{rs}w_r^\top w_s.  
\end{split}
\end{equation}

\begin{figure}[t]
\centering
\includegraphics[scale=0.28]{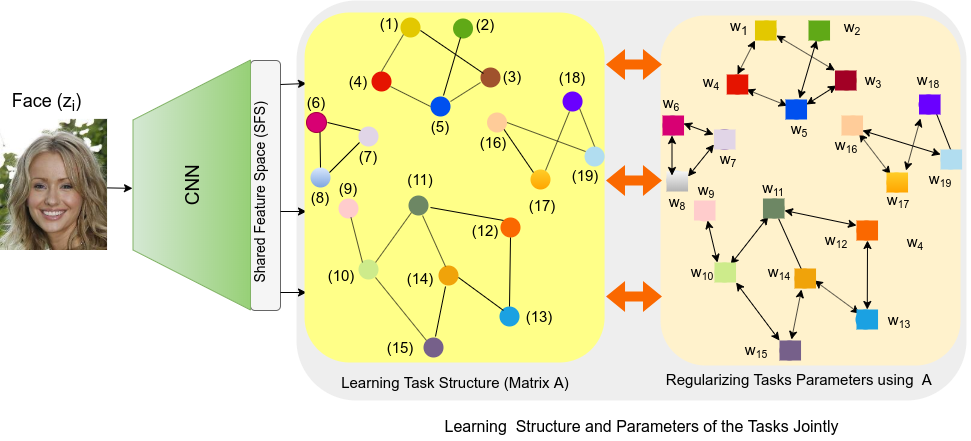}
\caption{Diagram of the second scenario: parameters and structure of the tasks are leaned jointly via Laplacian regularization framework.}
\label{arc_(1)}
\end{figure}
\begin{figure*}[t]
\centering
\includegraphics[scale=0.32]{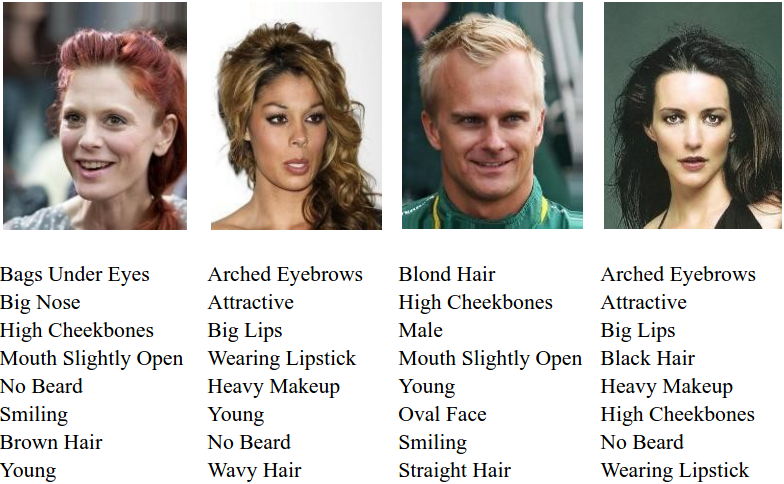}
\includegraphics[scale=0.31]{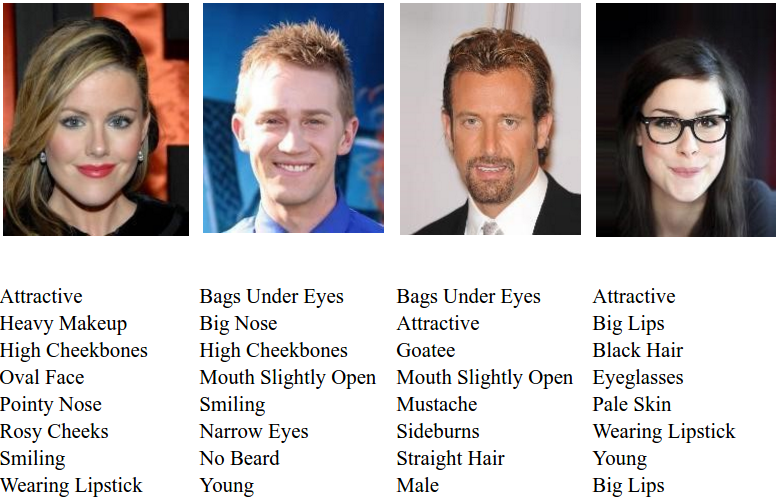}
\caption{Examples of face images used in our experiments from the CelebA and LFW datasets; each image is tagged by its facial attributes.}
\label{samples_}
\end{figure*}
\begin{algorithm}[t]
\caption{Learning parameters and graph of the tasks jointly}
\begin{algorithmic}[1]
\label{algo1}
\STATE {\textbf{input}:  TD: $(\mathcal{X},\mathcal{Y})=\{(x_i, y_i)\}_{i=1}^n$, $\xi$, $b$}
\WHILE{not converged}
\REPEAT
\STATE select a mini-batch:$(\mathcal{X}',\mathcal{Y}')=\{(x'_i, y'_i)\}_{i=1}^b$.
\STATE $W \leftarrow {\text{Optimize } \mathcal{L}(W, \mathcal{X'},\mathcal{Y'}) \text{ using SGD}}$
\UNTIL  for an epoch
\STATE $A \leftarrow \operatorname*{argmin} \{\mathrm{Tr}(WAW^\top)$ \\
\text{s.t. } $A \geq \epsilon I_{n\times n}, A_{off} \leq 0, A \textbf{1}_n = \epsilon \textbf{1}_n\}$
\ENDWHILE
\end{algorithmic}
\end{algorithm}
Here, $W_i$ is the \textit{i-th} column of     matrix $W$ created by staking parameters of the tasks column-wise.  Therefore, the loss function defined in (6) can be written as follows:

\begin{equation}
\mathcal{L}(\mathcal{W},\mathcal{X},\mathcal{Y})= min \sum_{i=1}^n \mathcal{L}_c(\mathcal{X},\mathcal{Y}_{i}, {w}_i)+ \xi \mathrm{Tr}(W A W^\top).  
\end{equation}
\begin{figure*}[t]
\centering
\includegraphics[scale=0.35]{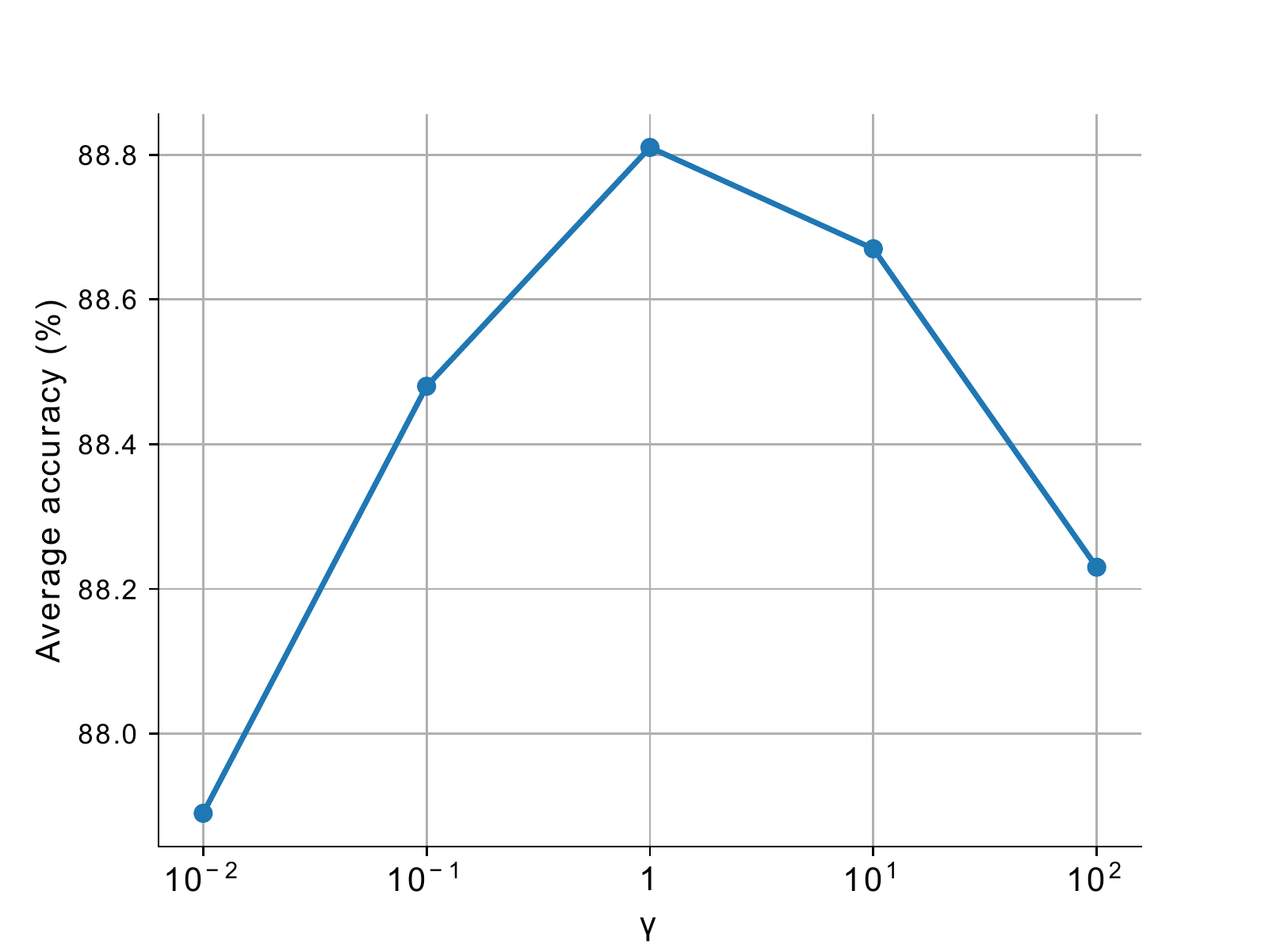}
\includegraphics[scale=0.35]{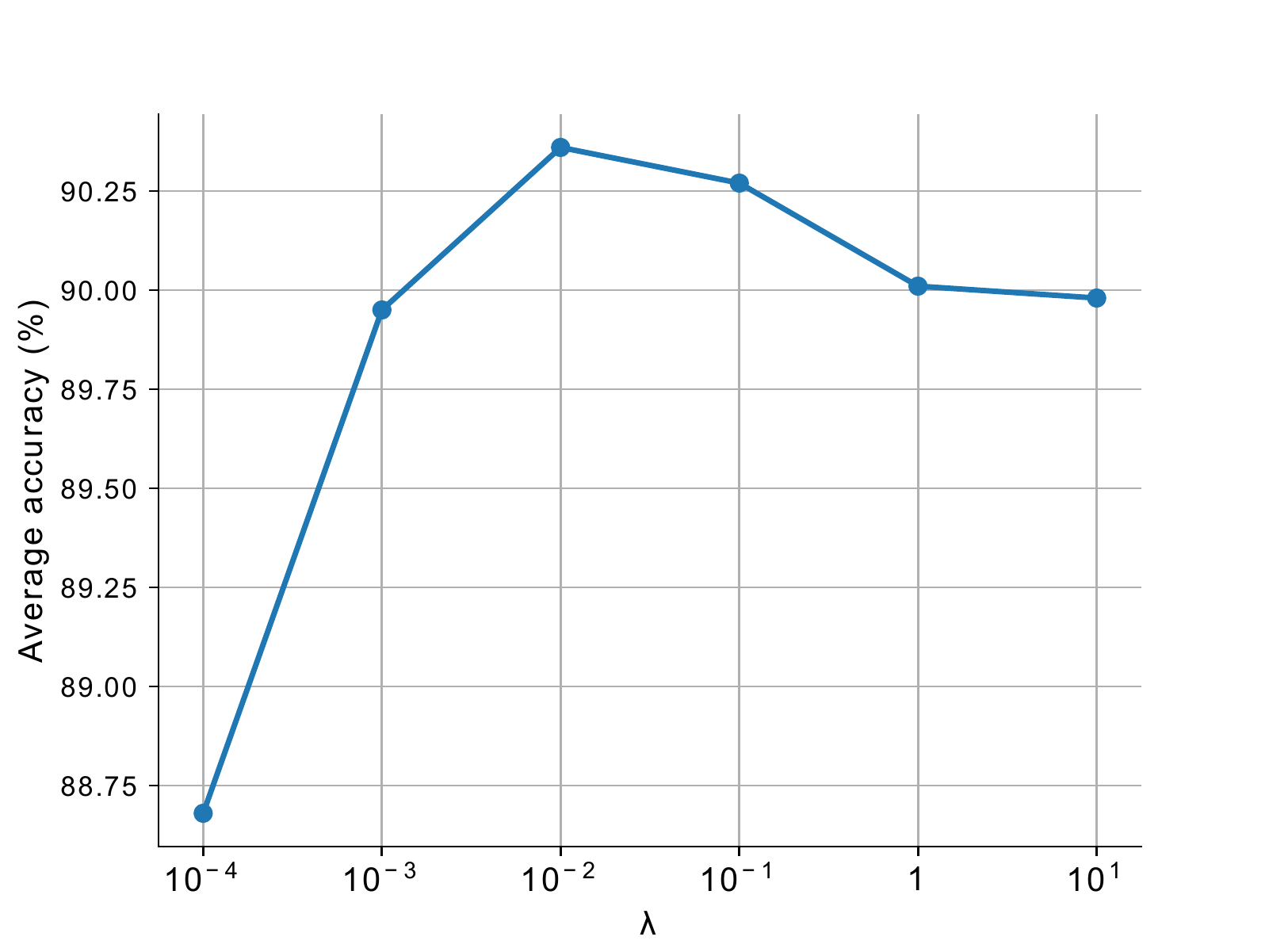}
\includegraphics[scale=0.35]{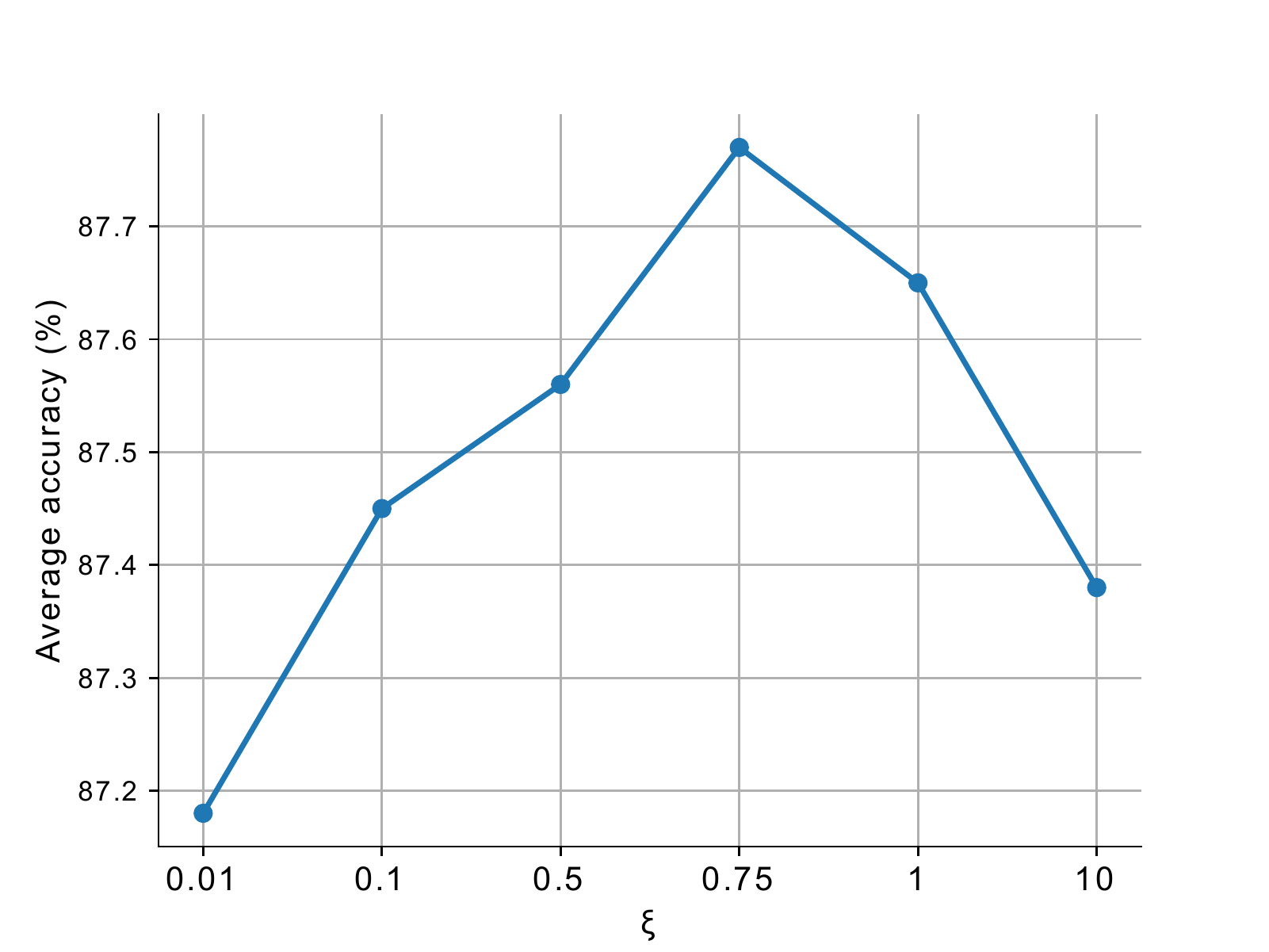}
\caption {Hyperparameters Tuning}
\label{hyp}
\end{figure*}
In this loss function, there are two unknown variables (i.e., the parameters of the tasks $W=[w_1,w_2,...,w_n]$ and the graph of tasks relatedness $A$), which we aim to learn them simultaneously.  Note that we can not choose any matrix $A$ in this optimization as a graph Laplacian. There are some constraints which ensure us the given matrix $A$ comes from a graph Laplacian as follows \cite{evgeniou2005learning,argyriou2013learning}: $\{A \geq \epsilon I_{n \times n}, A_{off} \leq 0, A \textbf{1}_n = \epsilon \textbf{1}_n\}$ where $A_{off}$ is  the off diagonal elements of the matrix $A$, and $\textbf{1}_n$ is a vector with size $n$ with elements equal to one. The optimization problem (14), given the constraints is not convex jointly in $A$ and $Q$, however, it is convex in one of two matrices (i.e, $W$ and $A$) when the other is fixed. Therefore, we can adopt an alternative minimization approach to optimize (14) presented in Algorithm 3.

Line 7 in Algorithm 3  is a Semi-Definite program (SDP) and can be solved by approaches such as interior point and the well-known packages such as Sedumi or SDPT3. Note that we stop this algorithm when there is no significant improvement on average accuracy of facial attribute predictors on the training data. Here, $\epsilon$ is  fixed and very small perturbation of the Laplacian penalty ($\approx  10^{-3}$). In case  $\epsilon$ is small, the graph term dominates the penalty and therefore, strong tasks similarities hold large weights in the graph and weak tasks similarities vice versa. Fig. \ref {arc_(1)} graphically represents that both the parameters and graph of the tasks are learned jointly. Here, each color denotes a specific task and the tasks are connected to each other based on their level of relatedness to create the graph. Next, parameters of the tasks are  pushed close to each other based on the graph via a Laplacian regularization.

\section{Experiment and Setup}

We evaluate the performance of our proposed model for multi-task facial attribute prediction. The outline of our study to evaluate the effectiveness of the proposed method is as follows: 1) We report performance of our facial attribute prediction model for both the cases where the structure of the tasks is known as a prior knowledge as well as the case where the structure of the tasks is unknown which we learn this structure over the course of training from our training data. As mentioned earlier in our methodology, we have followed the structure of the facial attribute predictors introduced in Liu et al \cite{liu2015deep}. 2) We compare our proposed model with other state-of-the-art methods for the facial attribute prediction problem. 3) We use the facial attributes for the application of face verification problem. Following the works  \cite{ranjan2017all, taherkhani2018deep, iranmanesh2018deep,hu2017attribute}  which have shown that incorporating facial attributes as a source of complimentary information potentially can improve the performance of the face verification models, we develop a Siamese network which takes advantage of the facial attributes in the verification loss function to improve the performance.



\subsection{Facial Attribute Benchmarks}

We applied our model to two standard datasets namely CelebA and LFW,  together which are annotated by 40 facial attributes. Examples of the images with their attributes are shown in Fig. \ref{samples_}

\textbf{CelebA}  is a large-scale face attribute and richly annotated dataset containing more than 200K celebrity images, each of which is annotated with  40 facial attributes.

\textbf{LFW} is a  well-known dataset for face recognition as well as attribute prediction. This dataset is helpful for addressing the problem of face recognition in an unconstrained environment.  The dataset contains more than 13k images of faces collected from the Internet, each of which is annotated with  40 attributes and five key points by a labeling company. CelebA and
LFWA include over eight million  and 500k attribute labels, respectively.
\begin{figure*}
\centering
\includegraphics[scale=0.271]{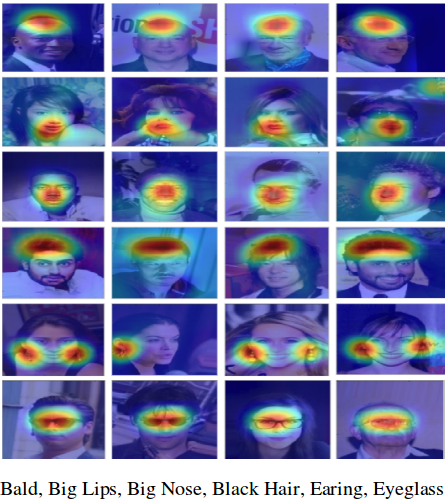}
\includegraphics[scale=0.271]{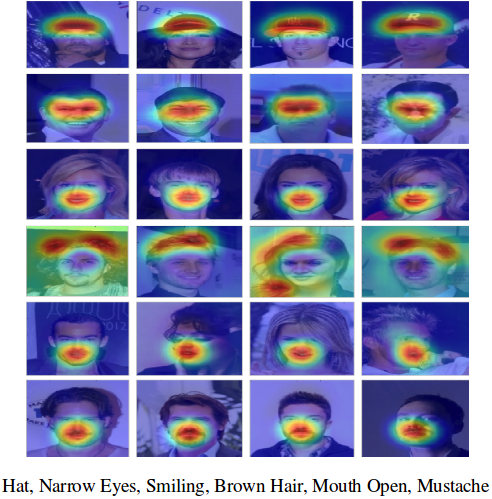}
\includegraphics[scale=0.271]{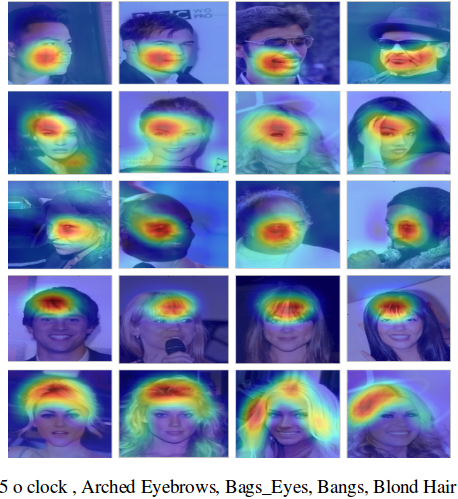}
\includegraphics[scale=0.271]{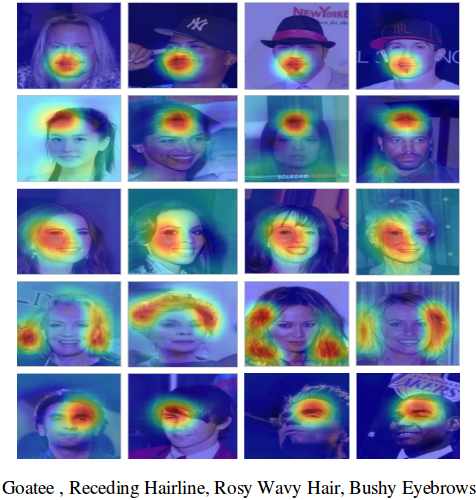}
\caption{CAM; the model uses GAP layer to localize the face region which highlights the corresponding attribute occurrence in the image. The figure shows exceptional ability of the GAP layer in facial attribute localization in our model. }
\label{cam2}
\end{figure*}
\begin{figure*}[h]
\centering
\includegraphics[scale=0.271]{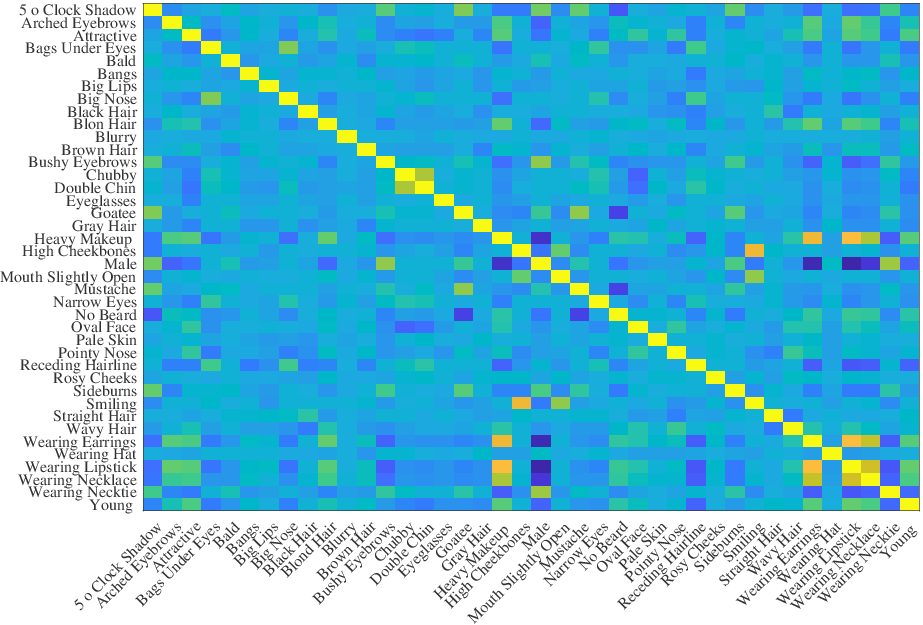}
\includegraphics[scale=0.271]{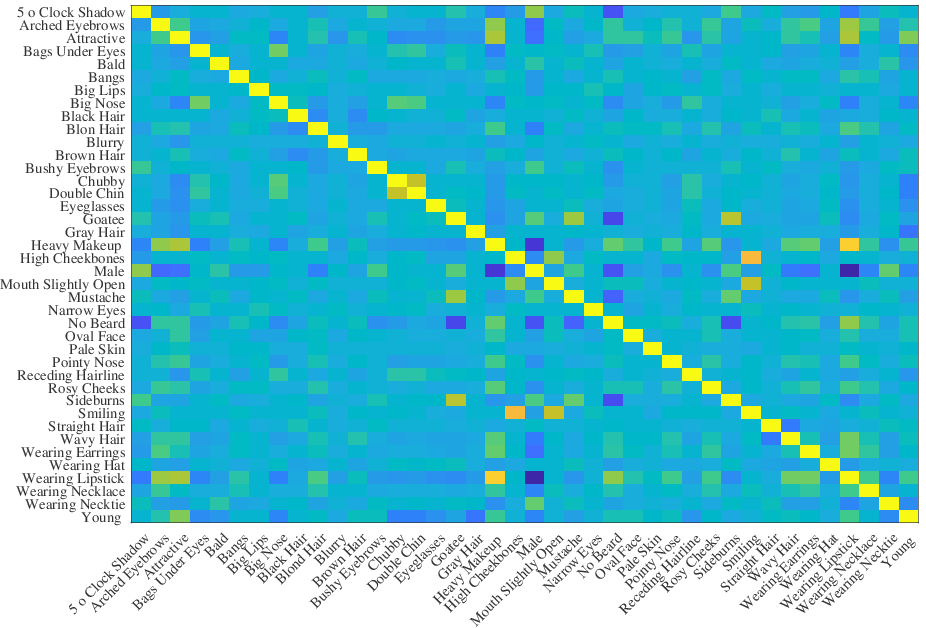}
\includegraphics[scale=0.135]{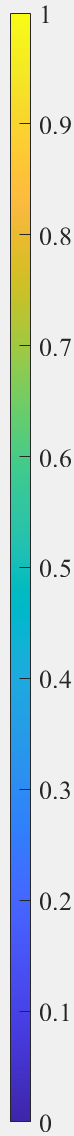}
\caption{Graph of tasks on the CelebA  (left) and LFW (right). Two attributes with stronger relationship have a larger value in the matrix.}
\label{samples_8}
\end{figure*}
\subsection{Hyperparameters Tuning} 
In this work, we use the VGG-16 architecture to extract the features from the face images. We use the SVD, and QR factorization implementation  of the Numpy library in the  Python in Algorithm 1. The batch size during the training is set to 128. We used the CelebA dataset to tune the hyperparameters of our model. CelebA is divided into three parts. Images of the
first eight thousand identities (160K images) are used as the  training set. The images of another one thousand identities (20K images)  are used as the validation set for the hyperparameters tuning. Then, we combine  the images of the first and the second part to train the entire model. The images from the rest of one thousand identities (20K images) are employed for testing the model. The hyperparameters of the model in the first scenario are $\lambda, \gamma$ in Eq (3), and hyperparameter of the model in the second scenario is $\xi$ in Eq (6). Fig. \ref{hyp} represents the average prediction accuracy for 40 facial attributes on the validation set of the CelebA dataset. Experimental results indicate that the best value for $\gamma$, $\lambda$, and $\xi$ are $1$, $10^{-2}$, and $0.75$, respectively. Moreover,  the results on hyperparameter tuning  in Fig. \ref{hyp} demonstrate that our proposed model is not notably sensitive to the hyperparameters.  


\subsection{Class Activation Map Using GAP Layer:} In order to show the power of SFS layer in image representation for our MT facial attribute prediction, we used a class activation map (CAM) which indicates that our model localizes discriminative image regions  to predict a particular facial attribute in the image. Here, attribute weights are provided  with
higher-order visual information which includes the spatial location of the attributes in the images. In other words, our model determines the regions of the image which are relevant to a particular facial attribute. In our model,  a global average pooling operation is performed on the feature maps obtained from the final convolutional layer, and then a linear  layer for each face attribute is conducted on the GAP outputs  to determine if such an attribute is present in the image. For example, in VGG-16, the last convolutional layer has 512 filters. For a $224\times 224$ input image, the output shape of the last convolutional layer is $512\times14\times 14$ due to applying 4 max-pooling operations from previous layers. For each of 512 channel, we have a $14 \times 14$ spatial mapping resolution. The GAP layer just takes each of these 512 channels and returns their spatial average. The parameters associated to a particular facial attribute predictor assigns a weight to each elements of the GAP layer output. For a given image, assume that $g_k(x,y)$ indicates the k-th feature map in the final convolutional layer at the spatial location $(x, y)$. For each of the facial attributes In CAM, the weights $w_k$ represent the significance of each of the channels in a way that the channels with high activation will have larger weights for localizing that attribute in the image. 
In order to obtain the class activation map, we plug average pooling $G_k=\sum_{x,y}g_k(x,y)$ into the class score $S_c$ which is a linear combination of GAP layer elements. Thus
\begin{equation}
  S_c=\sum_kw_k^c\sum_{x,y}g_k(x,y)=\sum_{x,y}\sum_k w_k^c g_k(x,y).  
\end{equation}
We declare $A_c$ as the CAM for facial attribute $c$ such that each spatial grid is determined as follows:
\begin{equation}
    A_c(x,y)=\sum_k w_k^c g_k(x,y).
\end{equation}

Therefore, we can write $S_c=\sum _{x,y}A_c(x,y)$ and in conclusion $A_c(x,y)$ directly represents the significance of the activation at each spatial element $ (x,y)$ which results in  prediction of the attribute $c$ in a given image. The CAM is directly obtained by a linear weighted sum of  visual patterns (i.e., $\{g_1,..,g_K\}$, where, $K$ is the number of feature maps) at different locations. 
Finally, we simply upsample the CAM to  the input image size by using nearest interpolation method. By performing this, we can recognize the image regions which are most related to a particular facial attribute.

\begin{table*}
\centering
\begin{tabular} {c|c|c|c|c|c|c|c|c|c|c|c|c|c|c}
\centering
& \rotatebox{90}{FaceTracer \cite{kumar2008facetracer}}& \rotatebox{90} {PANDA \cite{zhang2014panda}} & \rotatebox{90}{LNets+ANet \cite{liu2015deep}}  & \rotatebox{90}{MOON \cite{rudd2016moon} } &
\rotatebox{90}{NSA \cite{mahbub2018segment} } &
\rotatebox{90}{AUX \cite{hand2017attributes} } &
\rotatebox{90}{MCFA \cite{zhuang2018multi} } &
\rotatebox{90}{GNAS \cite{huang2018gnas}} &
\rotatebox{90}{PS-MCNN-LC \cite{cao2018partially}} &
\rotatebox{90}{DMM-CNN \cite{mao2020deep}} &
 \rotatebox{90}{MT-RBM-PCA \cite{ehrlich2016facial} }&
\rotatebox{90}{RCA \cite{sethi2018residual}} &
\rotatebox{90}{W/o prior knowledge} &\rotatebox{90}{Prior knowledge}
 \\
\hline
 5 o’clock Shadow & 85 & 88 &  91 & 94.03 & 93.13  & 94.51 &  94&  94.76  & 96.6  & 94.84 & 90 & 92.88 & 94.38    &  \textbf{98.33}\\ 
\hline
Arched Eyebrows &  76  &  78  &  79 & 82.26 & 82.56&  83.42 & 83 &  84.25 & 85.77 &   84.57 & 77 & 81.63  & 83.31  &\textbf{87.91}
 \\
 \hline
Attractive &  78 &  81  &  81  &  81.67  &  82.76 &   83.06 &    83 &   83.06 &   84.39 &   83.37  & 76 & 79.67 & 82.69 &   \textbf{85.98}\\
 \hline
Bags Under Eyes & 76  & 79  & 79  & 84.92 &  84.86 &  84.92  & 85 &  85.87  & 87.29 &  85.81 & 81 & 83.15 & 84.27 & \textbf{87.35} \\
\hline
Bald  & 89 &  96  & 98  &  98.77  & 98.03   & 98.9  &  99 & 98.96  &  99.41  & 99.03 & 98 & \textbf{99.52}  & 98.50&  98.97\\
\hline
Bangs & 88 & 92  & 95  & 95.8  & 95.71  &  96.05 &  96  & 96.2 & 98 &  96.22  & 88 & 94.51  & 96.26 & \textbf{98.77}
 \\
\hline
Big Lips & 64  & 67  & 68 &  71.48 &  69.28  & 71.47 & 72 &  71.79  & 73.13 &  72.93  & 69 &  79.89 & 71.09 & \textbf{74.86}
 \\
\hline
Big Nose & 74  & 75  & 78  & 84 &  83.81 &  84.53  & 84 &  85.1 &  86.4 & 84.78 & 81 & 83.67 & 84.75 &  \textbf{86.99}\\
\hline
Black Hair & 85  & 85  &  88  & 89.4  & 89.03  & 89.78  & 89   & 90.24 &  91.66  & 90.5 & 76 & 84.80  & 91.52 & \textbf{92.62}\\
\hline
Blond Hair & 93  & 93 &  95  & 95.86 &  95.76 &  96.01 &  96   & 96.11  & 97.93  &  96.13 & 91 & 94.97  & 95.71 & \textbf{98.83}\\
\hline
Blurry & 81 & 86  & 84 &  95.67  & 95.96  &  96.17  & 96 &  96.42 &  \textbf{98} & 96.4  & 95 & 96.57 & 96.89&  97.93\\
\hline
Brown Hair & 77 & 77  & 80  & 89.38 &  88.25 &  89.15 &  88 &  89.75  & 91.03 & 89.46 & 83 & 82.97 &  88.38 & \textbf{92.62}\\
\hline
Bushy Eyebrows & 86 & 86 &  90  & 92.62 &  92.66  & 92.84 &  92  & 92.99  & 94.51 &  93.01  & 88 & 91.36 & 92.37&  \textbf{95.15}\\
\hline
Chubby & 86 & 86  & 91  & 95.44 & 94.94 &  95.67  & 96  & 95.93  & 97.66  & 95.86  & 95 & 95.51 & 96.04 & \textbf{98.94} \\
\hline
Double Chin & 88 & 88  & 92 &  96.32 &  95.8 &  96.32  & 96   & 96.48 &  98.29 &  96.39 & 96 &  96.45  & 96.33&  \textbf{98.81}
 \\
\hline
Eyeglasses & 98 & 98  & 99 & 99.47  & 99.51 & 99.63 & \textbf{100}   & 99.69  & 99.85 &  99.69 & 96 & 98.18  & 98.83 &  98.87 
\\
\hline
Goatee & 93 & 93 &  95 &  97.04 &  96.68  & 97.24 &  97  & 97.59 &  97.74 &  97.63 &  96 &  96.77  & 96.95 & \textbf{98.85}
 \\
\hline
Gray Hair & 90  & 94   & 97   & 98.1   & 97.45 &  98.2   & 98  &  98.37  & \textbf{98.66 } &  98.27 & 97 & 97.92  & 97.59 & 97.64\\
\hline
Heavy Makeup & 85 &  90  &  90 &   90.99 &   91.59 &   91.55  &  92  &   91.82  &  93.31 &   91.85  & 85 & 89.72  & 91.15 &  \textbf{93.80}\\
\hline
High Cheekbones & 84 & 86   &  88   &  87.01   &   87.61  &   87.58    &  87  &   88.05  &   89.5  &  87.73 &83 & 86.74  & 88.95 &  \textbf{90.89} \\
\hline
Male & 91 &  97 &  98  &  98.1 &  97.95  & 98.17  & 98 &  98.5  & \textbf{98.81} &  98.29 & 90 & 95.87 & 97.72 &  98.55\\
\hline
Mouth Open & 87 &  93 &   92 &   93.54 &   93.78  &  93.74  &  93  &  94.16   &  95.99 &   94.16 & 82 & 89.81  & 93.87 & \textbf{96.79}\\
\hline
Mustache & 91 & 93  &  95 &   96.82 &   95.86 &   96.88 &   97  &  97.03 &   \textbf{98.56} &   97.03 &  97 & 96.31  & 95.1 &  97.48\\
\hline
Narrow Eyes & 82 & 84 &  81  &  86.52  &  86.88 &   87.23 &   87  &  87.66  &  89.07  &  87.73 & 86 & 90.64 & 89.14 &  \textbf{90.98}\\
\hline
No Beard & 90 & 93  &   95  &  95.58  &  96.17 &   96.05  &  96  &  96.3 &   98.03 &   96.41 & 90 & 94.64  & 96.68 & \textbf{98.47}\\
\hline
Oval Face & 64 & 65  &  66 &   75.73 &   74.93 &   75.84 &   75   &  75.57  &   77.43 &   75.89 & 73 & 76.55  & 76.16  & \textbf{78.93}\\
\hline
Pale Skin & 83 & 91  &  91 &   97 &   97 &   97.05 &   97   &  97.24 &   \textbf{98.84} &   97 & 96 & 96.93 &  97.49 & 97.56 \\
\hline
Pointy Nose & 68 &  71 &    72  &  76.46  &  76.47 &   77.47 &   77 &   78.24 &   79.32  &  77.19  & 73 & 76.95   & 78.25 &  \textbf{80.98}\\
\hline
Receding Hairline & 76 & 85  &   89  &  93.56 &   92.25 &   93.81  &  94 &   93.94 &   95.85 &   94.12 & 92 & 93.64  & 94.91 & \textbf{95.98}\\
\hline
Rosy Cheeks & 84 & 87  &  90  &  94.82 &   94.79 &   95.16  &  95  &  95.01 &  96.92   &  95.32 &94 & 95.32 & 96.95 & \textbf{97.79}\\
\hline
Sideburns & 94 & 93 &   96 &   97.59  &  97.17  &  97.85 &   98  &  97.96  &  \textbf{98.22} &   97.91 & 96 & 97.6  & 97.27 & 98.01\\
\hline
Smiling & 89 & 92  &  92  &  92.6  &  92.7 &   92.73  &  93  &  93.24 &   \textbf{94.85}  &  93.22 & 88 & 92.83  &  92.52 & 94.47
\\
\hline
Straight Hair & 69 & 69  &  73 &   82.26 &   80.41  &  83.58 &   85 &   84.77 &   85.96 &   84.72 & 80 & 81.19  & 85.18 & \textbf{86.71}\\
\hline
Wavy Hair & 73 & 77 &   80  &  82.47 &   81.7 &   83.91 &   85   &  84.52 &   86.39 &   86.01 & 72 & 75.42 & 84.4 & \textbf{86.75}\\
\hline
Earrings & 73 &  78 &   82 &   89.6 &   89.44 &   90.43 &   90   &  90.98 &   92.66 &   90.78 & 81 &  82.65 & 91.7 & \textbf{93.25} \\
\hline
Hat & 89 & 96  &  99  &  98.95 &   98.74  &  99.05  &  99  &  99.12 &   \textbf{99.43} &   99.12 & 97 & 97.93 & 98.34 & 98.41
 \\
\hline
Lipstick & 89 & 93  &  93  &  93.93 &   93.21 &   94.11 &   94  &  94.41 &   95.7  &  94.49 & 89 & 91.96 & 94.16 & \textbf{96.87}\\
\hline
Necklace & 68 & 67  &  71 &   87.04 &   85.61 &   86.63  &  88   &  87.61 &   88.98 &   88.03 & 87 &  \textbf{89.82} & 87.05 & 89.31 \\
\hline
Necktie & 86 & 91 &   93 &   96.63  &  96.05 &   96.51  &   97   &  96.76 &   \textbf{98.52} &   97.15 & 94 & 95.88 & 96.97 & 98.11 \\
\hline
Young & 80 & 84 &   87   &  88.08 &   88.01 &   88.48 &   88   &  88.89   &  90.54  &  88.98 & 81 & 86.63  & 89.03 &  \textbf{91.15}\\
\end{tabular}
\caption{Performance comparison of attribute prediction on CelebA dataset.}
\label{eval_table}
\end{table*}
\begin{table*}
\centering
\begin{tabular} {c|c|c|c|c|c|c|c|c|c|c}
\centering
& \rotatebox{90}{FaceTracer \cite{kumar2008facetracer}}& \rotatebox{90} {PANDA \cite{zhang2014panda}} & \rotatebox{90}{LNets+ANet \cite{liu2015deep}} & 
\rotatebox{90}{NSA \cite{mahbub2018segment} } &
\rotatebox{90}{AUX \cite{hand2017attributes} } &
\rotatebox{90}{MCFA \cite{zhuang2018multi} } &
\rotatebox{90}{PS-MCNN-LC \cite{cao2018partially}} &
\rotatebox{90}{DMM-CNN \cite{mao2020deep}} &
 \rotatebox{90}{W/o prior knowledge} &\rotatebox{90}{Prior knowledge}
 \\
\hline
 5 o’clock Shadow & 70  & 84 &  84  & 77.59  & 77.06 &  75  & 78.17  & 79.18 &79.25 &  \textbf{82.39 }\\ 
\hline
Arched Eyebrows &  67 & 79  & 82  & 81.72 &  81.78 &  79 &  83.53 &  82.7 & 84.04 & \textbf{87.67} \\
 \hline
Attractive &  71 & 81 & 83 &  80.16 &  80.31 &  77 &  81.84   & 81.1 & 83.52 & \textbf86.59{} \\
 \hline
Bags Under Eyes & 65  & 80 &  83 &  82.62 &  83.48 &  79  & 86.74  & 82.7 & 84.83 & \textbf{86.78} \\
\hline
Bald  & 77 & 84 &  88 &  91.88 &  91.94 &  91  & 92.6 &  91.96 & 93.12 & \textbf{95.35} \\
\hline
Bangs & 72 & 84  & 88  & 90.71 &  90.08 &  89 &  91.45 &  91.3 & 92.06 & \textbf{94.96}\\
\hline
Big Lips & 68  & 73 &  75 &  78.97  & 79.24  & 75  & 82.7  & 79.82 & 81.39 & \textbf{82.86}\\
\hline
Big Nose & 73  & 79  & 81 &  83.13 &  84.98  & 81  & 86.48  & 83.67 & 85.8 & \textbf{88.04 }\\
\hline
Black Hair & 76  & 87  & 90 &  92.49 &  92.63 &  91  & 92.96  & 91.55 & 93.85 & \textbf{94.89} \\
\hline
Blond Hair & 88  &  94  & 97  & 97.47  & 97.41 &  97 &  98.51  & 97.17
& 97.19 & \textbf{98.99} \\
\hline
Blurry & 73 & 74  & 74 &  86.42  & 85.23  & 86  & 87.2 &  87.58 & 90.07 &  \textbf{92.11} \\
\hline
Brown Hair & 62 & 74  & 77 &  80.93 &  80.85 &  77  & 81.87 &  81.56 & 81.70 &  \textbf{84.77}\\
\hline
Bushy Eyebrows & 67 & 79 &  82  & 84.26   & 84.97 &  76 &  85.72 &  85.33 & 86.78 & \textbf{88.11 }\\
\hline
Chubby & 67 &  69 &  73 &  76.06 &  76.86  & 74 &  78.11 &  77.66
 & 77.41 & \textbf{80.05} \\
\hline
Double Chin & 70 & 75 &  78 &  80.49 &  81.52  & 77  & \textbf{86.7}  & 80.98 & 83.07 & 85.1 \\
\hline
Eyeglasses & 90 & 89 &  \textbf{95} &  91.5 &  91.3 &  91 &  92.78 &   92.83 & 94.13  & 94.16 \\
\hline
Goatee & 69 & 75 &  78 &  83.01  & 82.97 &  80 &  84.11 &  82.82 & 83.42 & \textbf{86.29} \\
\hline
Gray Hair & 78 & 81  & 84 &  88.46  & 88.93  &  88 &  91.04  & 89.38 & 90.38 & \textbf{92.4}\\
\hline
Heavy Makeup & 88 & 93 &  95 &  95.39 &  95.85 &  94 &  96.6  & 95.68 &  96.95 & \textbf{98.12}\\
\hline
High Cheekbones & 77 &  86  & 88 &  88.34 &  88.38 &  85 &  88.77 &  88.13 & 90.19 & \textbf{91.38}\\
\hline
Male & 84 & 92 &  94  & 92.6 &  94.02 &  93  & 95.18  & 94.14 & 94.71 & \textbf{96.1}\\
\hline
Mouth Open & 77 & 78 &  82 & 82.5 &  83.51  & 78 &  84.6   & 84.45 & 84.35 & \textbf{86.08}\\
\hline
Mustache & 83 &  87 &  92 &  92.97 &  93.43 &  91 &  94.47 &  94.46 & 95.72 & \textbf{98.89}\\
\hline
Narrow Eyes & 73 & 73  & 81 &  82.75 &  82.86 &  78 &  83.51 &  83.67 & 84.70 & \textbf{87.88} \\
\hline
No Beard & 69 & 75 &  79 &  80.77 &  82.15 &  79 &  82.01 & 82.48 & 83.61 &  \textbf{86.29} \\
\hline
Oval Face & 66 & 72 &  74 &  76.8 &  77.39 &  74 &  77.9 & 76.94 & 80.10 & \textbf{82.84} \\
\hline
Pale Skin & 70 & 84 &  84 &  90.97 &  93.32 &  82 &  94.97  & 91.86 & 96.45 & \textbf{97.49}\\
\hline
Pointy Nose & 74 & 76  & 80 &  84.2 &  84.14 &  80  & \textbf{87.52}  & 84.51 & 86.82 & 87\\
\hline
Receding Hairline & 63 & 84  & 85 &  84.9 &  86.25 &  85  & \textbf{87.5} &  86.3 & 86.63 &  87.09\\
\hline
Rosy Cheeks & 70 & 73  &  78 &  87.08 &  87.92  & 85 &  88.81   & 86.44 & 89.52 & \textbf{92.65}\\
\hline
Sideburns & 71 & 76  & 77  & 81.76 &  83.13 &  78 &  84.42 &  82.99
& 83.93 &  \textbf{85.58}  \\
\hline
Smiling & 78 & 89 &  91  & 90.8 &  91.83  & 88 &  92.7  & 92.24 & 93.75 & \textbf{96.88} \\
\hline
Straight Hair & 67 & 73 &  76 &  78.91 &  78.53 &  77 &  79.65 &  79.2 & 82.45&  \textbf{84.51}\\
\hline
Wavy Hair & 62 & 75  & 76 &   78.28 &  81.61 &  79  & \textbf{83.35} &  79.87 & 81.97 & 83.05\\
\hline
Earrings & 88 & 92 &  94 &  94.75 &  94.95  & 93 &  95.54  & 94.14 & 96.09 & \textbf{98.92}\\
\hline
Hat & 75 & 82 &  88  & 90.23 &  90.07 &  91  & 91.21  & 90.84 & 92.15 & \textbf{93.17} \\
\hline
Lipstick & 87 &  93  & 95  &  94.07 &  95.04  & 94 &  95.7  & 95.11 & 96.28 & \textbf{98.19}\\
\hline
Necklace & 81 & 86 &  88 &  89.59 &  89.94 &  89 &  90.92  & 89.47 & 91.52 & \textbf{93.21}\\
\hline
Necktie & 71 & 79 &  79 &  81.4  & 80.66  & 82  & 82.18  & 81.28 & 83.70 & \textbf{85.13}\\
\hline
Young & 80 & 82  & 86  & 85.68 &  85.84  & 87 &  86.88 &  88.94 & 87.37 & \textbf{89.94}\\
\end{tabular}
\caption{Performance comparison of attribute prediction on LFW dataset.}
\label{eval_table_1}
\end{table*}
\begin{figure*}[t]
\centering
\includegraphics[scale=0.435]{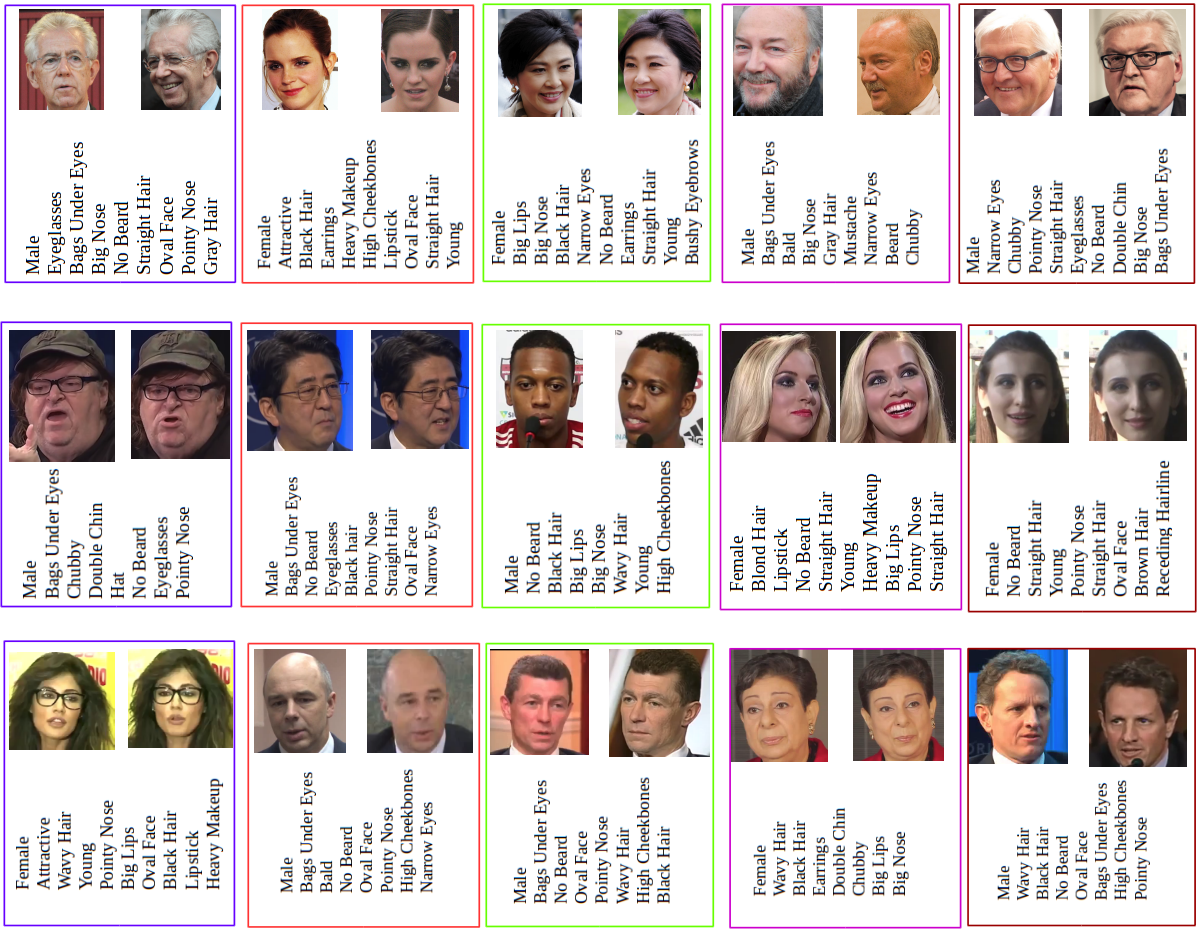}
\caption{Examples of facial attribute prediction using our model on the IJB-A, IJB-B, and IJB-C datasets from up to down, respectively.}
\label{samples_11}
\end{figure*}
\subsection{Further Analysis on  the Facial Attribute Graph }
We intuitively know that all the facial attributes occurring in the image are not related to each other. For example, “hair color” is not related to the “gender” or some other facial attributes such as "smiling". Although the weight between two unrelated attributes in the  graph is a small value, we still remove such relationships in the  graph by putting a simple threshold on the values of  the matrix $A$ (see Eq. 14) in a way that all the elements of the matrix  smaller than a threshold are set to zero. In our experiments we set this threshold to 0.1.  The facial attribute graph  as shown in Fig. \ref{threl_44} indicates that " mouth slightly open, high cheekbones , smiling", "attractive, no beard, heavy makeup, wavy hair, young, bangs, brown hair, oval face, pointy nose, rosy cheeks, wearing lipstick, blonde hair, wearing necklace, arched eyebrows, wearing earring" , " wearing necktie, male, mustache, sideburns, bald, 5 o clock shadow, goatee, bushy eyebrows", "wearing hat " , "gray hair", " blurry", "pale skin", "eyeglasses", "black hair, straight hair" and  "  receding hairline, chubby, double chin, bags under eyes, big lips, big nose, narrow eyes" are clustered in the same group, respectively. However, for each pair of facial attributes from different groups, there is no coupling and interaction in the  graph. The original matrices obtained from the training on the CelebA and LFW datasets are demonstrated in Fig. \ref{samples_8}
\begin{figure*}
\centering
\includegraphics[scale=0.3695]{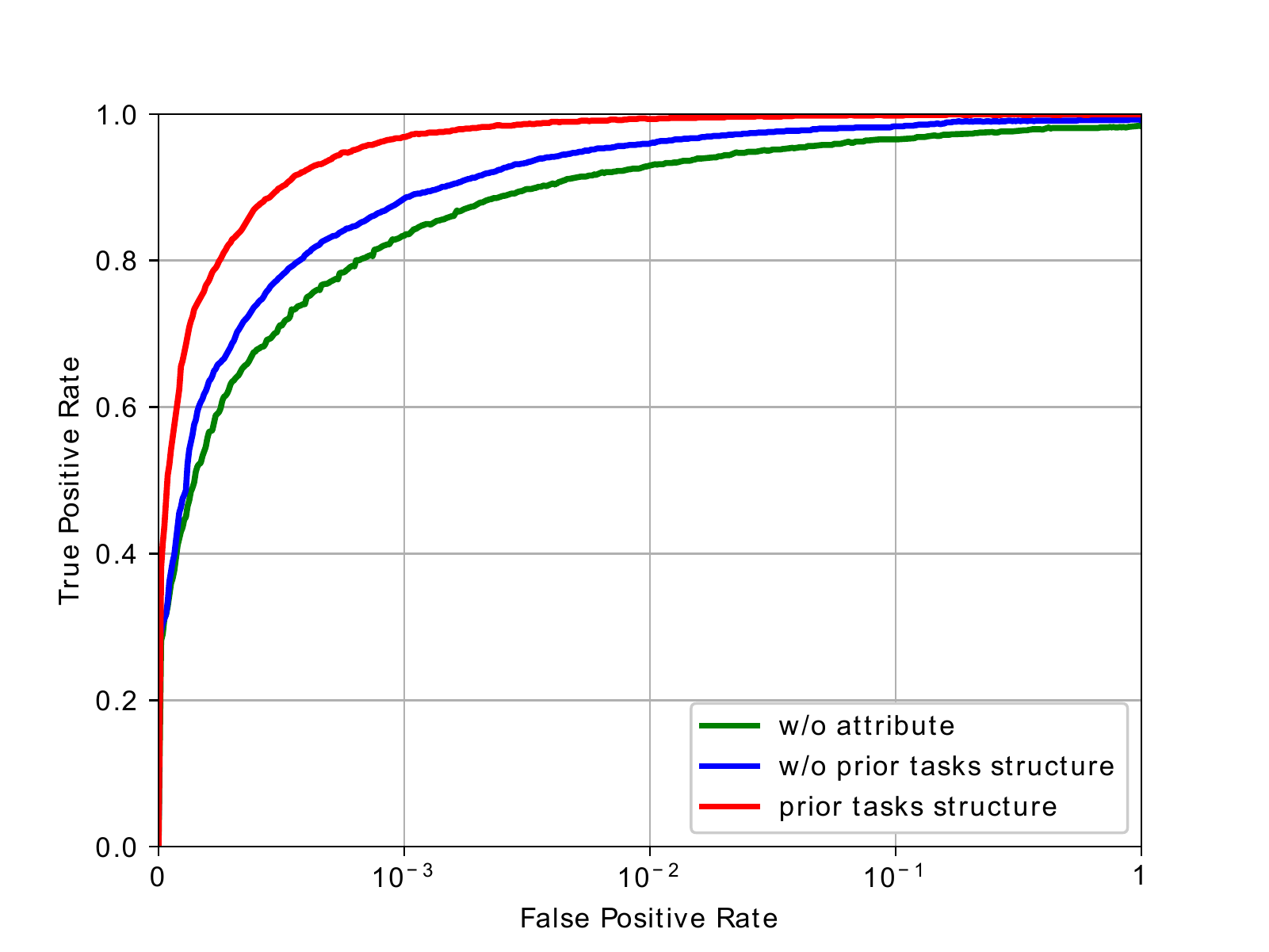}
\includegraphics[scale=0.3695]{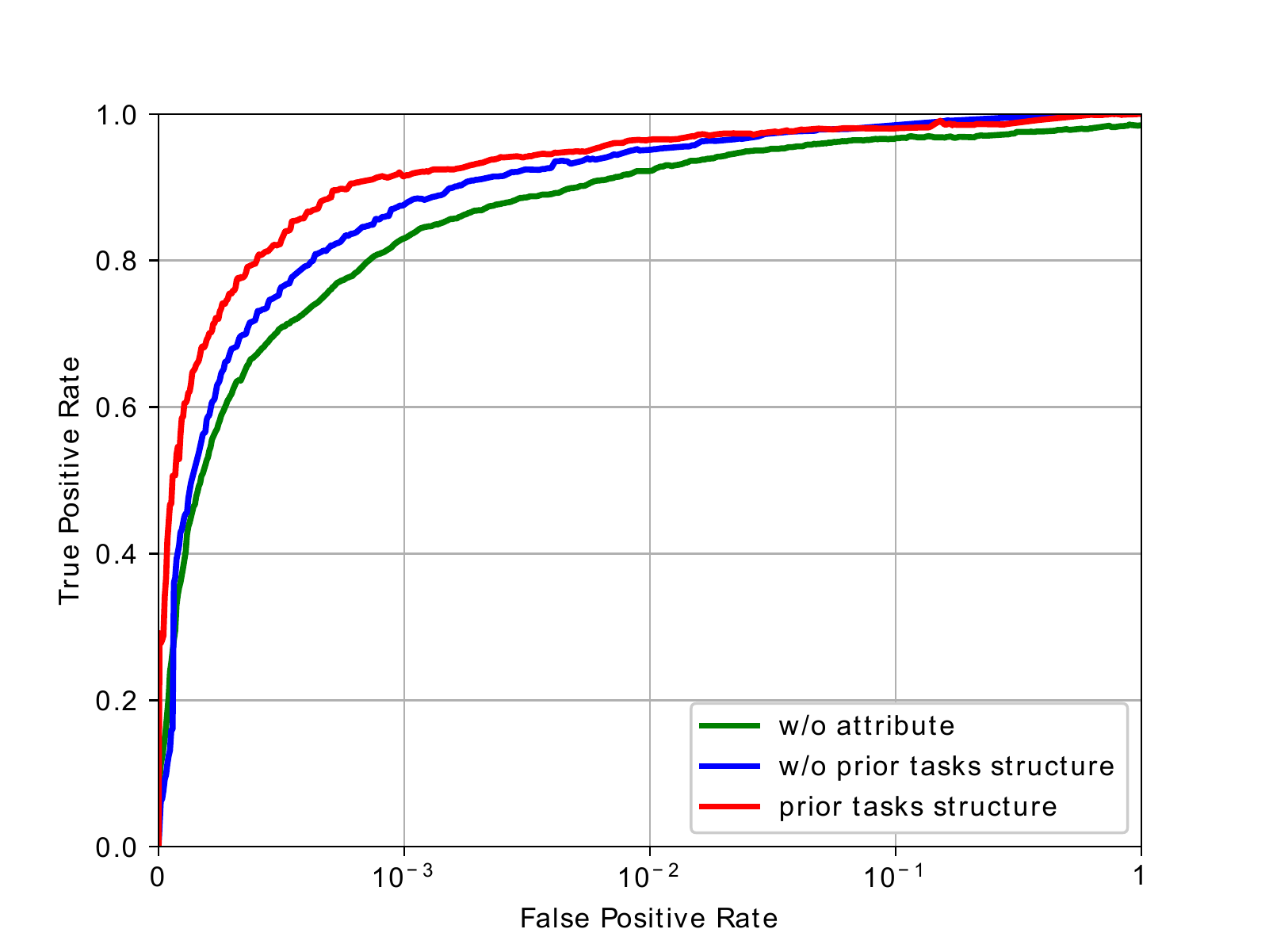}
\includegraphics[scale=0.3695]{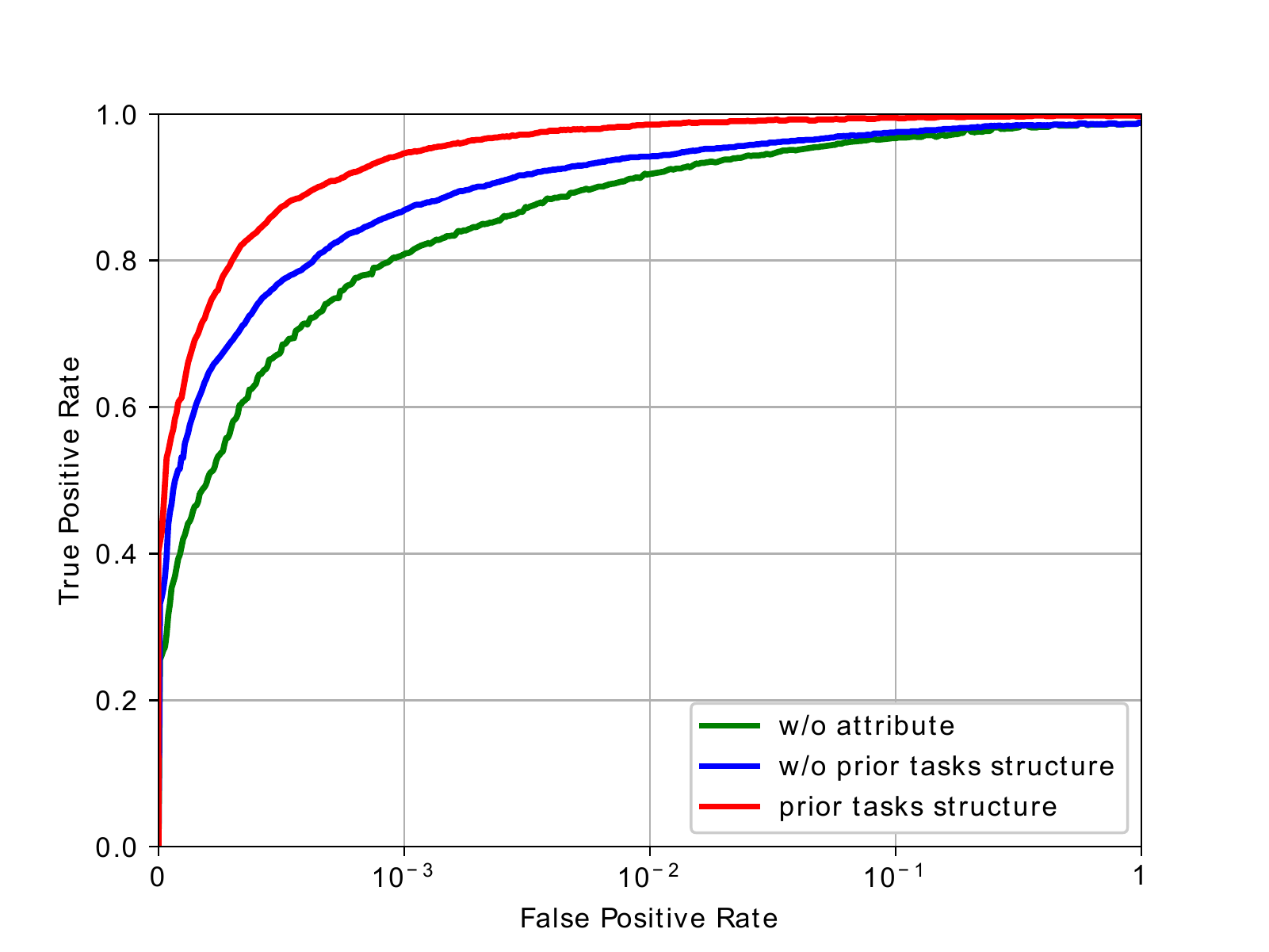}

\caption{ROC curves on the IJB-A, IJB-B, and IJB-C datasets from left to right respectively.}
\label{cam2_2}
\end{figure*}
\begin{figure}[t]
\centering
\includegraphics[scale=0.24]{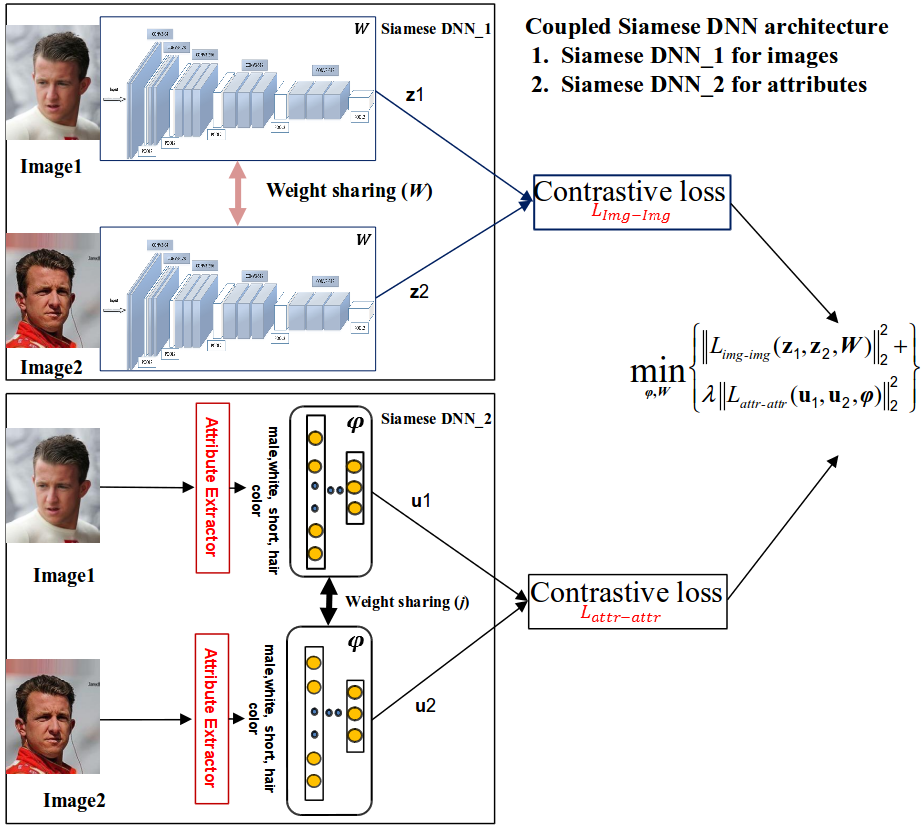}
\caption{Face verifier assisted by facial attributes.}
\label{thr_}
\end{figure}

\subsection{Methods for Comparisons}
We compare our method with  several competitive approaches \cite{mahbub2018segment,hand2017attributes,zhuang2018multi,huang2018gnas,cao2018partially,mao2020deep} including well-known facial attribute prediction methods such as FaceTracer \cite{kumar2008facetracer}, PANDA \cite{zhang2014panda}, ANet+ LNet \cite{liu2015deep}, MT-RBM-PCA \cite{ehrlich2016facial}, MOON \cite{rudd2016moon} and RCA \cite{sethi2018residual} in Table \ref{eval_table} \& Table \ref{eval_table_1}. All the methods presented in this paper use half of  the LFWA  dataset. Approximately  6, 263 images are used for training and the remaining images are used for testing. We divided CelebA into three different partitions. For the ANet+LNet method, images of the first eight thousand identities, which is roughly  162k images, are employed for pre-training and fine-tuning. The images of the next one thousand identities,  which is roughly twenty thousand  images, are used to train the classifier. And the remaining one thousand identities, which is nearly twenty thousand  images,  are used for testing. For fair comparison in our experiment, we have been consistent with the train and test split of these datasets as used in other methods. We have used half of LFW dataset for training and  remaining half for testing. For CelebA we have used images of eight thousand identities for training and images of the remaining one thousand identities for testing. The compared results in Table  \ref{eval_table} and Table \ref{eval_table_1} show that our method in case  where there is no prior knowledge (i.e., W/o prior knowledge in Tables) on the structure of the attributes is still comparable with the state of the art methods, not only for global attributes (attributes which are predicted from the whole region of the face such as "chubby", "male", "young" and "attractive") but also for fine-grained facial attributes such as  "mustache", "pointy nose" and "narrow eyes". Moreover, our results indicate that our model in case where there is prior knowledge on the structure of the attributes (i.e., prior knowledge in Tables) outperforms the sate-of-the art methods for the most of the global and fine-grained attributes.

\subsection{Application on Face Verification}

The IARPA Janus Benchmark A (IJB-A) \cite{IJB-A} is a  challenging dataset collected under complete unconstrained conditions. IJB-A contains 500 subjects with 5,712 images and 20,414 frames extracted from videos. Following the standard protocol in \cite{IJB-A}, we evaluate our method on  verification task.  The IARPA Janus Benchmark C (IJB-C) dataset \cite{IJB-C} builds on IJB-A, and IJB-B \cite{IJB-B} datasets and has a total of 31,334 images for a total number of 3,531 subjects. We have also evaluated our method on IJB-A and IJB-C datasets. Here,  we  perform  face  verification  using  FaceNet pre-trained on the VGGFace2 dataset  \cite{cao2018vggface2}. VGGFace2 is a large-scale face recognition dataset, where the images are downloaded from Google Image Search and have large variations in pose, age, illumination, and ethnicity. The dataset contains about 3.3 million images corresponding to more than 9000 identities with an average of 364 images per subject. 

Here, since these datasets are not annotated by facial attributes, we initially predict facial attributes using our approach based on both  prior knowledge and without prior knowledge, and then follow our previous work \cite{talreja2019attribute} which uses  a Siamese-based network integrated by facial attributes to perform the face verification task. The examples of facial attribute prediction using our model on the IJB-A, IJB-B, and IJB-C datasets are indicated in Fig. \ref{samples_11}. Moreover, the Siamese-based network which uses facial attributes for face verification is indicated in Fig. \ref{thr_}. The ROC curves on the IJB-A, IJB-B, and IJB-C datasets in Fig. \ref{cam2_2} indicate that our facial attribute predictor models have positive influence on improving of the face verification task. The results also indicate that the better facial attribute model provides more positive contribution on the performance of the face verifier. This is because, in our experiments in Fig. \ref{cam2_2}, the results show that our model with given prior knowledge (red ROC curves) outperforms the case where the  prior knowledge is not available during the training.

Here, we briefly explain our face verifier.  The final objective of our face verifier as shown in Fig. \ref{cam2_2} is to find the global deep latent features in a common embedding subspace representing the relationship between the genuine and imposter pairs. To find this common subspace, we couple the Siamese network via a contrastive loss function $L_{cont}$ \cite{Chopra2005LearningAS}. This loss function  $(L_{cont})$ is minimized so as to drive the genuine pairs towards each other in a common embedding subspace, and at the same time, push the impostor pairs away from each other. Let $x^i$ denote the input face image. $c(i,j)$ is a binary label, which is equal to 0 if $x^i$ and $x^j$ belong to the same class (i.e., genuine pair), and equal to 1 if $x^i$ and $x^j$ belong to the different class (i.e., impostor pair). Let $z_1(.)$ and $z_2(.)$ denote the deep convolutional neural network (CNN)-based embedding functions to transform  $x^i$ and $x^j$, respectively into a common latent embedding subspace. Then, contrastive loss function $(L_{cont})$  if $c(i,j)=0$ (i.e., genuine pair) is given as:
\begin{equation}
\begin{split}
L_{cont}(z_1(x^i),z_2(x^j),c(i&,j)) = \\ & 
  \frac{1}{2}\left\lVert z_1(x^i)-z_2(x^j)\right\rVert^2_2.  
  \end{split}
  \end{equation}
Similarly if $c(i,j)=1$ (i.e., impostor pair), then contrastive loss function $(L_{cont})$ is :
  \begin{equation}
  \begin{split}
L_{cont}(z_1(x^i),&z_2(x^j),c(i,j))  = \\ & \frac{1}{2}\mbox{max}\biggl(0,m-\left\lVert z_1(x^i)-z_2(x^j)\right\rVert^2_2\biggr), 
\end{split}
\end{equation}
where $m$ is the contrastive margin and is used to "tighten" the constraint. Therefore, the total loss function for coupling the sub-networks is denoted by $L_{cpl}$ and is given as: 



\vspace{-0.25cm}
\begin{equation}
    \begin{split}
        L_{cpl}=\frac{1}{N^2}\sum_{i=1}^{N}\sum_{j=1}^{N}L_{cont}(z_1(x^i),z_2(x^j),c(i,j)), 
    \end{split}\label{eq:6}
\end{equation}
where $N$ is the number of training samples. The main motivation for using the coupling loss is that it has the capacity to find the discriminative embedding subspace because it uses the class labels implicitly, which may not be the case with some other metric such as Euclidean distance. This discriminative embedding subspace would be useful for matching of the face images. Similar to the contrastive loss for face verification, another important objective of our face verifier is the matching of the facial attributes using the  contrastive loss. In other words, the identity attributes such as gender which are invariant are encouraged to be identical for genuine pairs while being different for the imposter pairs.   However, separating these two objectives by learning multiple CNNs individually is not optimal since different objectives may share common features and have hidden relationship, which can be leveraged to jointly optimize the objectives. This notion of joint optimization has been used in \cite{Zhong2016FaceAP}, where they train a CNN for face recognition, and utilize the features for attribute prediction. Therefore, for this task, we use the respective feature set from the common embedding subspace to also match the attributes for the given images. Also, our network shares a large portion of its parameters among different attribute prediction tasks in order to enhance the performance of the verification task in a mutli-task learning paradigm.  Comparing the ROC curves in Fig. \ref{cam2_2} demonstrate that using attributes in our Siamese-based face verifier (blue and red curves) improve the baseline verifier where the attributes are not used (green curves).

\section{Conclusion}
In this paper,  we consider two scenarios for multi-task facial attributes
prediction. In the first scenario, the structure of the tasks is given as
a prior knowledge during the training. In this scenario,
 the level of dependency among facial attributes has not
been specified, but the clustering pattern of the facial attributes 
are  provided as a prior knowledge for optimizing the parameters of
the attribute predictors in a multi-task learning paradigm. The inductive bias used in this scenario is that the facial attributes within the same cluster lie in a low dimensional subspace and parameters of the predictors within the same group are represented by a linear combination of a limited number of underlying basis tasks. Here, a sparsity constraint on the coefficients of this linear combination is also considered such that each task is represented in a more structured and simpler fashion. In the second scenario, however, the structure of the tasks is unknown
and we tend to learn the parameters and structure of the tasks
jointly using a Laplacian regularization framework based on the kernel methods. Here, we formulate facial attribute prediction as an optimization problem in a reproducing kernel Hilbert space, in which the kernel are
also learned. Here, we compared the performance of our facial attribute prediction model for both the cases where the structure of the tasks is known as a prior knowledge as well as the case where the structure of the tasks is unknown which we learn this structure over the course of training from our training data. The results indicate that our multi-task facial attribute prediction model in the first case outperforms the the second case which means that the prior knowledge provides useful information for the facial attribute prediction in our multi-task learning paradigm. Then, we compare our proposed model with other state-of-the-art methods for the facial attribute prediction problem. Our experiments on two facial attribute  datasets  indicate that our multi-task CNN attributes predictors are comparable with the state of the art for both global and fine-grain facial attributes. Finally,  we used our proposed facial attribute prediction models for the application of face verification. Here, we used a Siamese-based network which takes advantage of the facial attributes in the verification loss function to improve the performance.

{\small
\bibliographystyle{IEEEtran}
\bibliography{egbib}
}
\end{document}